%% file: main.tex
\documentclass[twoside]{article}
\usepackage[accepted]{aistats2025}

\PassOptionsToPackage{authoryear}{natbib}

\usepackage{multirow}
\usepackage{array}
\usepackage{booktabs}
\usepackage{graphicx}  % Required for including images
\usepackage{subcaption}  
\usepackage{paralist}
\usepackage{natbib}

\usepackage[utf8]{inputenc} % allow utf-8 input
\usepackage[T1]{fontenc}    % use 8-bit T1 fonts
\usepackage{hyperref}       % hyperlinks
\usepackage{url}            % simple URL typesetting
\usepackage{booktabs}       % professional-quality tables
\usepackage{multirow}
\usepackage{amsthm}
\usepackage{amsfonts}       % blackboard math symbols
\usepackage{nicefrac}       % compact symbols for 1/2, etc.
\usepackage{microtype}      % microtypography
\usepackage{xcolor}         % colors
\usepackage{bm}

\newtheoremstyle{exampleStyle} % Name of style
  {3pt} % Space above
  {3pt} % Space below
  {}    % Body font (empty for normal text)
  {}    % Indent amount
  {\bfseries} % Theorem head font
  {.}   % Punctuation after theorem head
  { }   % Space after theorem head
  {}    % Theorem head spec

\theoremstyle{exampleStyle}
\newtheorem{example}{Example}

\usepackage{graphicx}

\usepackage{tabularray}
\UseTblrLibrary{booktabs}
\usepackage{algorithm}
\usepackage{algpseudocode}

\input{math_commands.tex}

\usepackage{amsmath}
\usepackage{amssymb}
\usepackage{mathtools}
\usepackage{amsthm}
\usepackage{float}
\usepackage{pifont}% http://ctan.org/pkg/pifont
\usepackage{booktabs}

\newtheorem{proposition}{Proposition}

\newtheorem{lemma}{Lemma}
\usepackage[textsize=tiny]{todonotes}
\begin{document}

\twocolumn[

% \aistatstitle{Joint learning of conditional diffusions and summary networks for neural posterior estimation}

\aistatstitle{Conditional diffusions for amortized neural posterior estimation}

\aistatsauthor{ Tianyu Chen{$^\dagger$}\And Vansh Bansal{$^\dagger$}  \And  James G. Scott$^{\dagger,*}$ }

\aistatsaddress{$^\dagger$Department of Statistics and Data Sciences, University of Texas at Austin \\
$^*$McCombs School of Business, University of Texas at Austin}]

% \title{Conditional diffusions for neural posterior estimation}

% % The \author macro works with any number of authors. There are two commands
% % used to separate the names and addresses of multiple authors: \And and \AND.
% %
% % Using \And between authors leaves it to LaTeX to determine where to break the
% % lines. Using \AND forces a line break at that point. So, if LaTeX puts 3 of 4
% % authors names on the first line, and the last on the second line, try using
% % \AND instead of \And before the third author name.

% \author{
%   Tianyu Chen, Vansh Bansal, and James G. Scott \\
%   University of Texas at Austin
% }

\begin{abstract}

Neural posterior estimation (NPE), a simulation-based computational approach for Bayesian inference, has shown great success in approximating complex posterior distributions. Existing NPE methods typically rely on normalizing flows, which approximate a distribution by composing many simple, invertible transformations. But flow-based models, while state of the art for NPE, are known to suffer from several limitations, including training instability and sharp trade-offs between representational power and computational cost. In this work, we demonstrate the effectiveness of conditional diffusions coupled with high-capacity summary networks for amortized NPE. Conditional diffusions address many of the challenges faced by flow-based methods.  Our results show that, across a highly varied suite of benchmarking problems for NPE architectures, diffusions offer improved stability, superior accuracy, and faster training times, even with simpler, shallower models. Building on prior work on diffusions for NPE, we show that these gains persist across a variety of different summary network architectures. Code is available at \url{https://github.com/TianyuCodings/cDiff}.
\end{abstract}

\section{Introduction}

Neural posterior estimation (NPE), a type of simulation-based inference (SBI), has gained significant traction in recent years for approximating complex posterior distributions $p(\theta \mid X)$, particularly in scenarios where the likelihood function is either intractable or conceptualized as a ``black box'' recipe for simulating the data-generating process. NPE methods for Bayesian inference \cite[e.g.][]{geffner2023compositional,sharrock2022sequential,wildberger2023flow,deistler2022truncated,radev2020bayesflowlearningcomplexstochastic,greenberg2019automatic} have proven to be remarkably effective in a wide range of scientific applications, from cosmology \cite{leyde2024gravitational} to epidemiology \cite{anirudh2022accurate}, where traditional computational methods for likelihood-based inference are computationally prohibitive or impractical.  NPE methods can also be effective in more conventional---but still challenging---statistical inference settings, where the likelihood is available but the posterior remains difficult to sample.

\paragraph{Diffusions vs. normalizing flows.} The success of NPE has built upon the power of generative models, with normalizing flows \cite[e.g.,][]{radev2020bayesflowlearningcomplexstochastic} emerging as the dominant approach over the last several years. However, flow-based models exhibit significant limitations. In particular, they encounter a fundamental trade-off between expressiveness and tractability: accurately capturing complex posterior distributions often necessitates either deep architectures—composing numerous simple invertible flows—or more sophisticated flows with Jacobians that are computationally challenging to evaluate. Both choices lead to substantial memory and computational overhead. Moreover, when applied to NPE, normalizing flows frequently suffer from instability and, in some cases, divergence during training—a phenomenon that we systematically observe across multiple examples.

% Compared to normalizing flows, diffusion models constitute a more expressive class of generative models and have demonstrated remarkable success in fields such as image generation \cite{ho2020denoising,karras2020analyzing}. Furthermore, diffusion models exhibit superior sample efficiency relative to normalizing flows. Under mild assumptions, recent results on sample complexity for diffusion models, such as Lemma 1.4 in \cite{gupta2024improved}, establish that $\tilde{O}(d/\epsilon^2)$ samples are required to obtain a good approximation, where $d$ is the dimension of the distribution and $\epsilon$ denotes the score estimation error. In contrast, for normalizing flows, Theorem 23 in \cite{li2024sampling} presents a negative result, demonstrating that normalizing flows require $\tilde{O}(\epsilon^{-d})$ samples to achieve a good approximation, per Definition 16 of \cite{li2024sampling}. This lower bound on the error indicates the best possible approximation error attainable given a fixed number of samples. Crucially, since the dimension $d$ appears in the exponent, normalizing flows struggle to avoid the curse of dimensionality.

Compared to normalizing flows, diffusion models constitute a more expressive class of generative models and have demonstrated remarkable success in fields such as image generation \cite{ho2020denoising,karras2020analyzing}. Furthermore, diffusion models have been shown to exhibit favorable sample efficiency relative to normalizing flows. Recent results suggest that, under mild assumptions, diffusion models can approximate target distributions with a sample complexity that scales quite favorably in the dimension $d$ of the random variable being modeled \cite{gupta2024improved}. In contrast, theoretical analyses indicate that normalizing flows may suffer from sample complexity scaling that can become prohibitive---even exponential in $d$---in high-dimensional settings, making accurate approximation increasingly difficult \cite{li2024sampling}. This challenge arises due to fundamental limitations in how normalizing flows represent complex distributions, leading to difficulties in avoiding the curse of dimensionality.

There is a very recent line of work on diffusions for NPE \cite[e.g.,][]{geffner2023compositional,sharrock2022sequential}, but only in the non-amortized case. Specifically, prior work in this direction eschews the use of a summary network, instead assuming that all observations \( x_1, x_2, \dots, x_n \) are known \textit{a priori} and treated as a fixed-size conditioning input to the diffusion model. This stands in contrast to the main body of NPE work on normalizing flows, which tends to incorporate summary networks inspired by DeepSets \cite{bloem2020probabilistic}, LSTMs \cite{hochreiter1997long}, or set attention mechanisms \cite{lee2019set}.  While the ``no summary network'' assumption enables a decoder to be tailored to a specific \( x \), it comes at the cost of needing to retrain the whole network for each new dataset \( \{x_i\}_{i=1}^{n} \). Moreover, there is a specific issue that arises in conditional diffusions without summary networks: with $n$ data points, one must sum score estimates over all \( n \) observations \citep{sharrock2022sequential}, thereby leading to a higher-variance score. This leads to potential instability in the sampling process, making posterior inference less reliable. Additionally, when \( x \) is high-dimensional while \( \theta \) remains low-dimensional, the lack of a summary network results in inefficient training. In such cases, the network expends substantial capacity processing \( x \), while the gradient flow remains insufficient for learning an accurate posterior over \( \theta \). This ultimately limits inference quality.

%as an alternative to normalizing flows for NPE.  To be sure, diffusions have been widely used as generative models in the ML literature, but they have not been comprehensively studied as a possible NPE solution. In Bayesian inference, the goal is not just to generate plausible data samples, but to provide accurate uncertainty quantification for the parameters of a known "forward" model (i.e. likelihood) $p(X \mid \theta)$ that is assumed to have generated the data. In this paradigm, posterior distributions are typically lower-dimensional, but often have much higher uncertainty (i.e. entropy), compared to the distributions encountered in, say, image generation. Moreover, getting the uncertainty right is crucial. Scientists rely on precise characterization of posterior distributions to draw valid inferences about unknown parameters, conduct hypothesis tests, or make decisions under uncertainty. In this setting, distinct posterior modes represent distinct scientific explanations for the observed data, and so mode collapse is especially problematic. More generally, a failure to capture the full range of uncertainty in the posterior not only limits predictive accuracy but can also mislead decision-makers in fields where NPE has proven valuable, such as epidemiology and climate science.  NPE methods must excel in capturing the full shape of the posterior, including tail behavior, multimodality, and sharp transitions in probability. The potential benefits of diffusions are poorly understood in this context.  

\begin{figure*}[t!] % Use [t] to place the figure at the top of the page
    \centering
    \begin{subfigure}[c]{0.30\textwidth} % Set subfigure width to 32% of the text width
        \includegraphics[width=\textwidth]{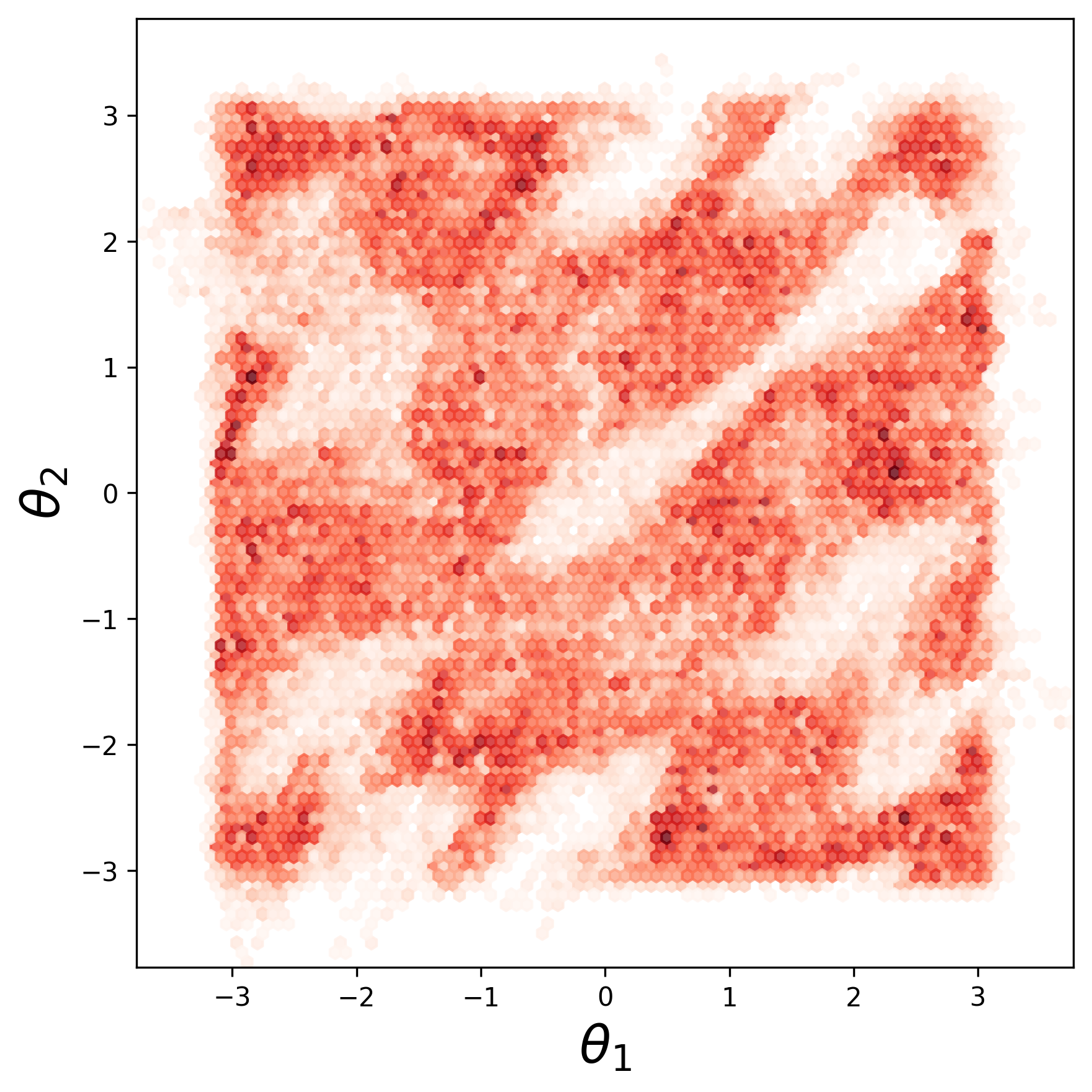}
        \caption{cNF sampling}
        \label{fig:figure1}
    \end{subfigure}
    \hfill
    \begin{subfigure}[c]{0.30\textwidth} % Set subfigure width to 32% of the text width
        \includegraphics[width=\textwidth]{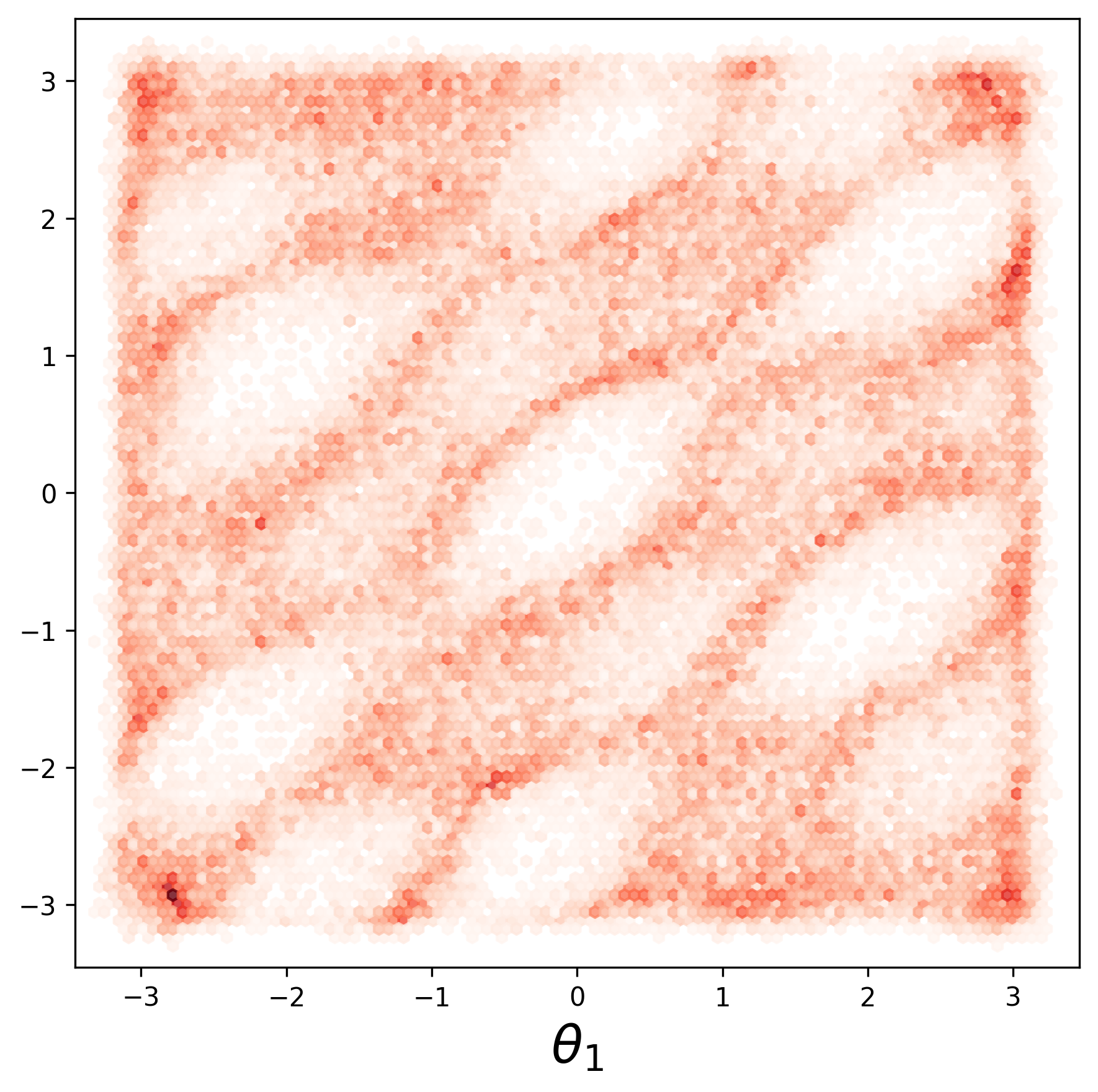}
        \caption{cDiff sampling}
        \label{fig:figure2}
    \end{subfigure}
    \hfill
    \begin{subfigure}[c]{0.33\textwidth} % Set subfigure width to 32% of the text width
        \includegraphics[width=\textwidth]{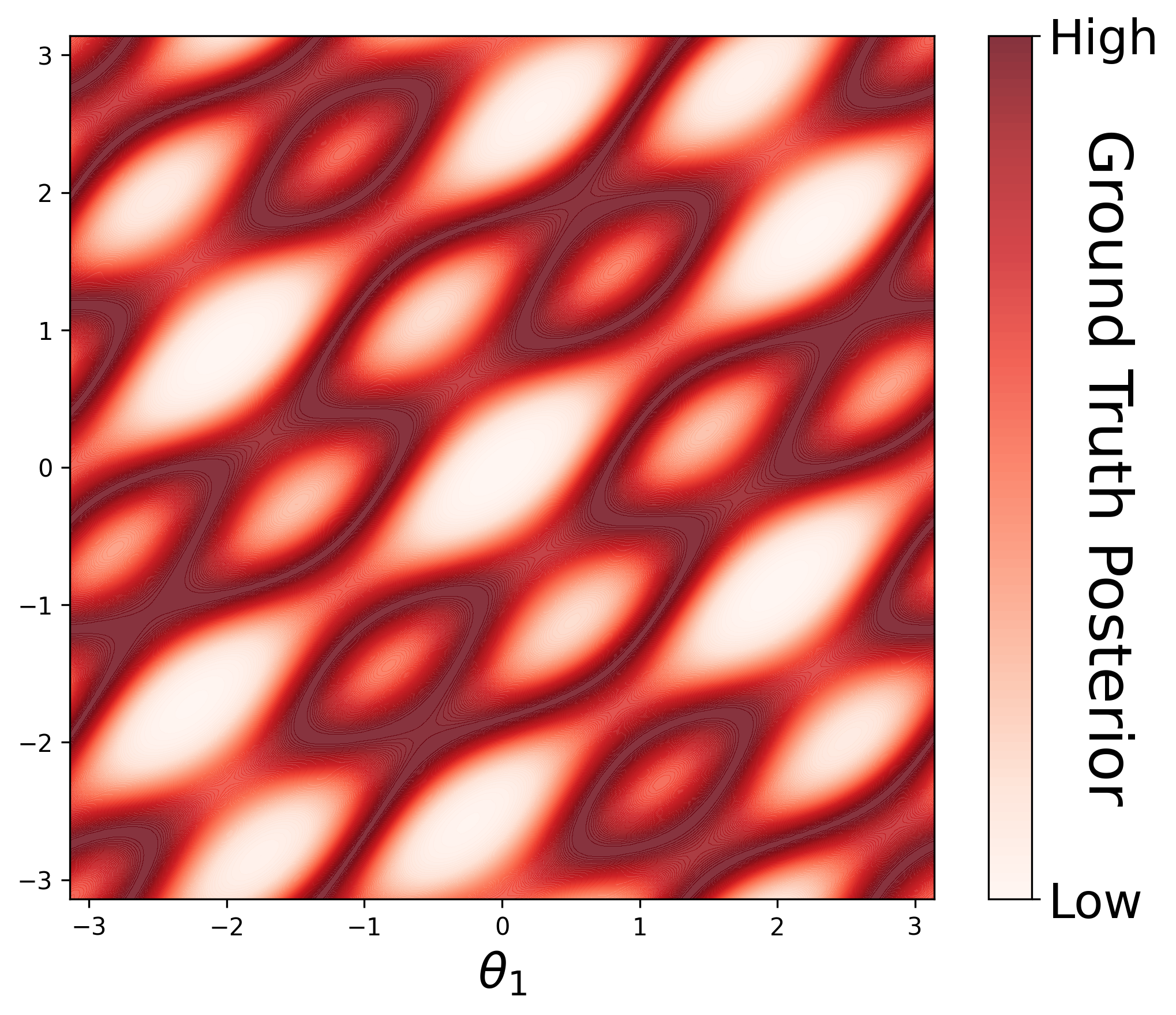}
        \caption{Ground Truth}
        \label{fig:figure3}
    \end{subfigure}
    \caption{Cosine example (above). We generate $10^5$ samples from both fitted posterior approximations (normalizing flows and diffusions), choosing $y=0$ for the observed data. The diffusion decoder is visibly better at capturing the undulating, multimodal character of the true posterior, a result confirmed by our experiments in Section \ref{sec:results}.}
    \label{fig:cosine_example}

\vspace{3mm}

    \begin{subfigure}[b]{0.94\columnwidth} % Set subfigure width to fit within a single column
        \includegraphics[width=0.95\textwidth]{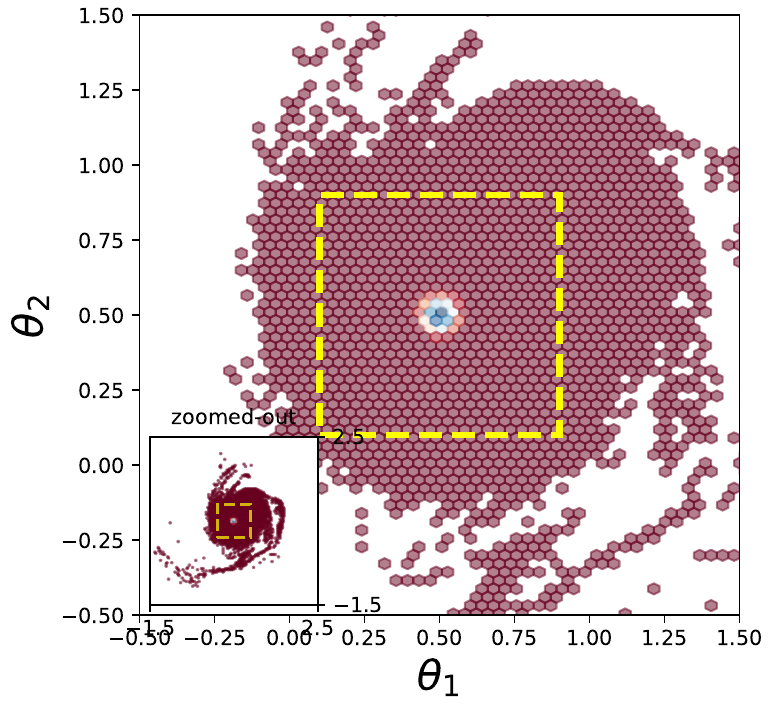}
        \caption{cNF sampling}
    \end{subfigure}
    \begin{subfigure}[b]{0.94\columnwidth} % Set subfigure width to fit within a single column
        \includegraphics[width=1.15\textwidth]{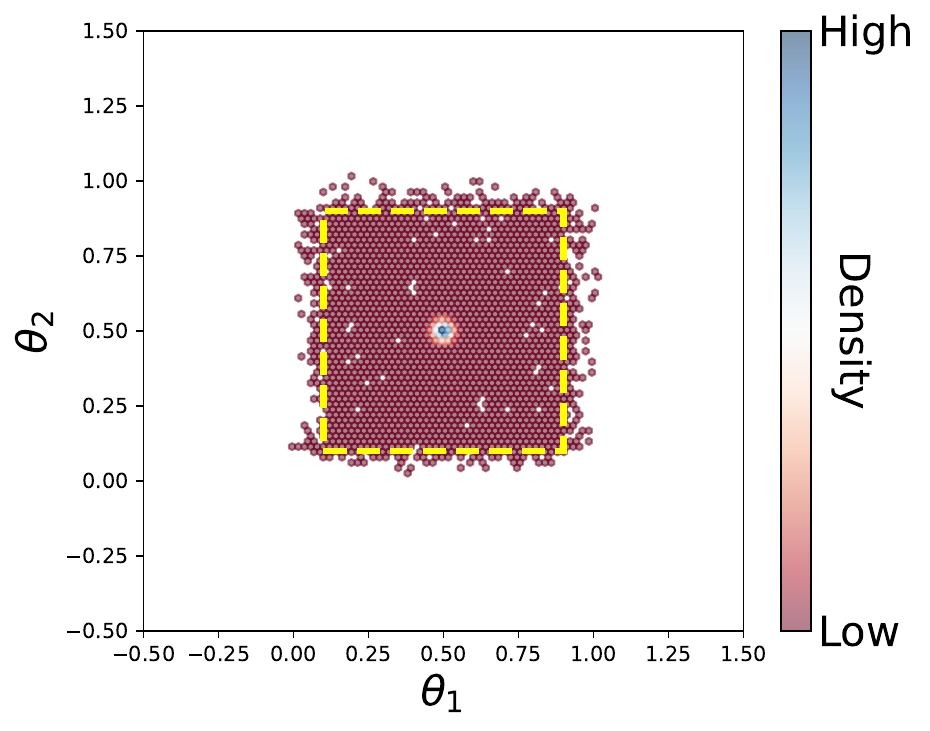}
        \caption{cDiff sampling}
    \end{subfigure}
    \caption{Witch's hat example (above). Draws from the estimated posteriors are shown for normalizing flows (left) versus diffusions (right). Here the observed data is $y=(0.5,0.5)$. The prior for $\theta$ is uniform on $[0.1,0.9]^2$, shown as a bounding box, but neither model was given explicit knowledge of the posterior's bounded support. The diffusion model is much better at characterizing the sharp transition between the region of uniform nonzero probability inside $[0.1,0.9]^2$, to the region of zero probability immediately outside it.}
    \label{fig:witch_hat_example}
    
\end{figure*}

\paragraph{Summary of Contributions.}

On the theoretical side, we give a formal justification for training a diffusion decoder jointly with a summary network that extracts a learned fixed-dimensional representation of an arbitrarily sized data set. Specifically, our main result shows that jointly training the summary network and diffusion decoder gives a valid upper bound on the Kullback-Leibler divergence between the true posterior and the estimated posterior.  This addresses addresses a clear gap in the NPE literature: summary networks aren't considered in prior work on conditional diffusions, yet they are an essential component of modern NPE architectures, especially in the amortized case.  

On the empirical side, we introduce a benchmark suite for NPE architectures that extends beyond existing benchmarks by incorporating a diverse range of statistical inference problems. Notably, our suite includes scenarios where the likelihood function is known but posterior sampling remains challenging. These ``white-box likelihood'' problems address a gap in prior work, enabling a more comprehensive evaluation of NPE methods on problems familiar to statisticians. Using this benchmark suite, we demonstrate that conditional diffusion models, when employed as NPE decoders, offer several key advantages over flow-based methods. These advantages include enhanced stability and improved accuracy. Notably, these benefits hold across both exchangeable and sequential data settings and persist regardless of the specific encoder or summary network architecture used.

\paragraph{Three short examples.}

Before detailing our approach and results, we first highlight three short examples that illustrate the superior performance of conditional diffusions versus normalizing flows for approximating complex posterior distributions. These examples are included in our benchmarking suite, but our presentation here is not intended to be exhaustive or formal. Rather it is meant to provide intuition and illustrate key aspects of our results, using simple default settings for each decoder. We use cNF as the abbreviation for the conditional Normalizing Flow model and cDiff for the conditional Diffusion model.

A key aspect of these examples is that, when designing the NPE architectures, we deliberately ignore certain ``obvious'' posterior features, such as boundary constraints or non-identified parameters. The goal is to evaluate how well these architectures can handle challenging posterior structures without being explicitly tailored for them. This approach provides insight into the flexibility of each architecture in managing complex posteriors (e.g.~an unknown boundary constraint as an example of a sharp transition in probability), especially where important structural features are unknown and the architecture cannot be specifically adapted, e.g. via transformations that remove constraints.

\begin{example}[Sum of cosines]

Our first example is intended as a toy version of a system with coupled oscillators; it also bears similarity to problems involving beamforming in phased array antennas. Let \( \theta = (\theta_1, \theta_2) \) have a uniform prior on \( (-\pi, \pi)^2 \). Given \( \theta \), the observed data \( y \) is sampled from \( y \mid \theta \sim N(f(\theta), 1) \), where \( f(\theta) \) is a sum of three cosines of different frequencies and phases, introducing nontrivial correlation and multiple modes into the posterior:
\[
f(\theta_1, \theta_2) = \cos(\theta_1 - \theta_2) + \cos(2 \theta_1 + \theta_2) + \cos(3 \theta_1 - 4 \theta_2) \, .
\]
We trained a diffusion model and a normalizing flow to approximate the posterior $p(\theta \mid y)$ for this model; the normalizing flow was necessarily deeper (32 layers vs. 4), but for the sake of a fair comparison, the models were constructed to have very similar numbers of parameters (204,338 parameters for diffusion model and 215,296 parameters for the normalizing flow), and each saw the same amount of training data, enough to yield apparent convergence of both models. We then queried each fitted model with the test point $y_{\text{obs}} = 0$ and drew 100,000 posterior samples. Because the prior is uniform on $(-\pi, \pi)^2$, the posterior is proportional to the likelihood, $p(\theta \mid y_{\text{obs}} = 0) \propto \exp\left(-0.5 \cdot (f(\theta))^2\right) $.  

In Figure \ref{fig:cosine_example}, the normalizing flow clearly struggles, producing a posterior that misses many clear modes and troughs. The diffusion model, while far from perfect, much more faithfully renders the undulating ridges of the true posterior. One could likely get better performance with more complex flows; the same could likely be said of a more complex diffusion model. The point here is simply that one gets decent ``out of the box'' performance with a relatively simple diffusion decoder, but that, at best, getting good performance from a flow-based decoder would require substantially more effort.  

\end{example}

\begin{figure*}[t]
\centering
\begin{subfigure}[b]{0.9\textwidth} % Adjust the width of each subfigure
        \includegraphics[width=\textwidth]{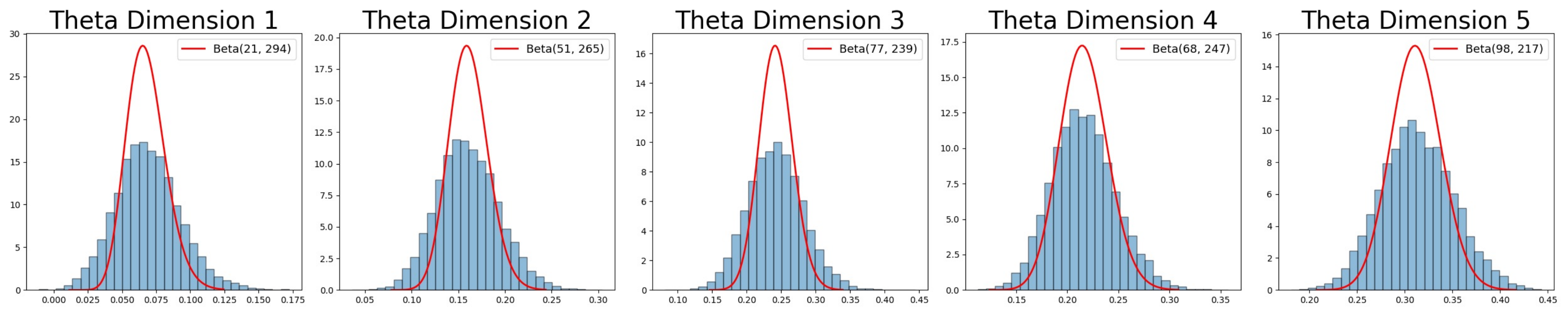}
        \caption{Draws from the normalizing flow approximation (above).}
\end{subfigure}
\vfill
\begin{subfigure}[b]{0.9\textwidth} % Adjust the width of each subfigure
        \includegraphics[width=\textwidth]{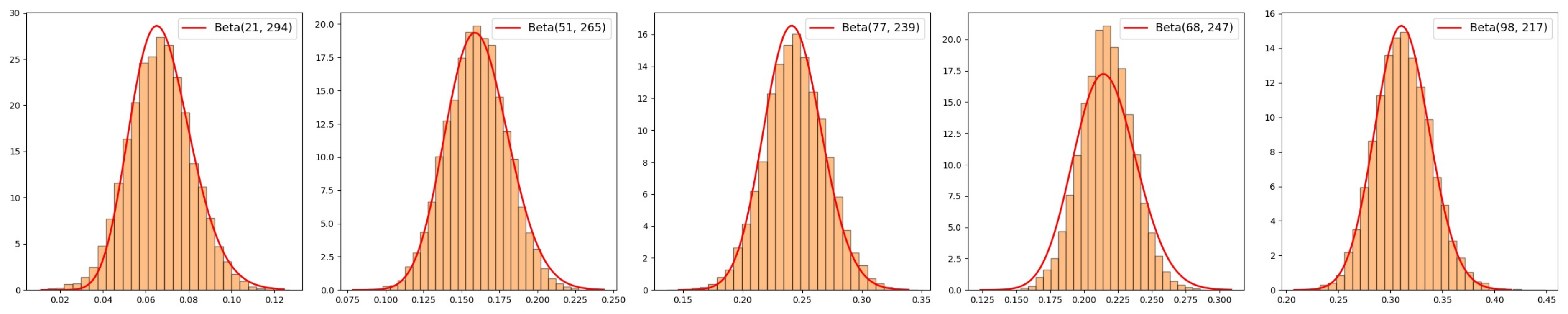}
        \caption{Draws from the conditional diffusion approximation (above).}
\end{subfigure}
    \caption{\label{fig:dirichlet_example} Dirichlet-multinomial example. We show the marginal of each method's estimated posterior for $\theta$ based on $N=315$ and $\hat{p}=(0.0667, 0.1600, 0.2367, 0.2200, 0.3167)$. Since the true posterior is a conjugate Dirichlet distribution whose marginals are beta distributions, the ground truth marginal (red line) is shown for reference.}
\end{figure*}

\begin{example}[Witch's hat]
The witch's hat distribution is a famous example of a posterior distribution that exhibits poor MCMC mixing \cite{matthews1993_witchhat}.  Let the prior for \( \theta \in \mathcal{R}^P \) be uniform on $[0.1, 0.9]^P$, a subset of the unit hypercube that excludes the region near the boundary. Given \( \theta \), the  data \( y \in \mathbb{R}^P \) follows a mixture distribution:
\[
(y \mid \theta) \sim \delta \cdot U([0, 1]^P) + (1 - \delta) \cdot N(\theta, \sigma^2 I),
\]
where $\delta$ and $\sigma$ are both small; the \( N(\theta, \sigma^2 I) \) component is the high conical peak of the witch's hat, and \( U([0, 1]^P) \) is the broad, flat brim.  This distribution was constructed by \cite{matthews1993_witchhat} as an example where the mixing time of the Gibbs sampler increases exponentially with dimension; our modified version introduces the extra wrinkle that the parameter space has a region of zero probability not shared by the sample space, since $\theta \in [0.1, 0.9]^P$.  We trained both a normalizing flow and a diffusion to estimate this posterior, using the same settings as in the previous example. Because the prior is flat on $[0.1, 0.9]^P$, the posterior based on a single sample $y$ is proportional to the witch's hat likelihood, truncated to the support of the prior.  Thus the posterior exhibits two challenging transitions: from the sharply peaked (but log-concave) Gaussian region to the flat-but-nonzero region inside $[0.1, 0.9]^P$, and again from the flat-but-nonzero region to the region of zero probability outside $[0.1, 0.9]^P$. Neither model was given explicit knowledge of this hard boundary.

Figure \ref{fig:witch_hat_example} shows the results for the toy example where $P=2, \sigma = 0.02, \delta=0.05$. The normalizing flow performs substantially worse at characterizing the sharp transitions in the posterior, especially the jump at the boundary of $[0.1, 0.9]^P$---a problem becomes substantially worse with increasing dimension $P$.
\end{example}

\begin{example}[Dirichlet-multinomial]
Our third example is a canonical model in Bayesian inference: the conjugate Dirichlet-multinomial model. We expect any decoder to do reasonably well on this simple problem, but we include it because the parameter space (the probability simplex) is closed and bounded, while the support of the base (Gaussian) distribution of both decoders is $\mathcal{R}^P$. As with Example 1, neither decoder is given explicit knowledge of the bounds of the parameter space.  Our goal was to see how well both decoders perform on a problem that seems simple, but where a perfect approximation is actually impossible, because the base and target distributions have fundamentally different topological structures.

Figure \ref{fig:dirichlet_example} shows both reconstructions on a five-dimensional problem.  Both model are reasonably good at capturing the point estimates and getting the uncertainty (i.e. posterior dispersion) approximately right. But the diffusion is visibly better, a performance advantage borne out by our benchmarks in Section \ref{sec:results}.

\end{example}

\section{Preliminaries and related work}
\label{sec:preliminaries}
Suppose we observe data from a known probabilistic model $p(X \mid \theta)$, where $\theta \in \Theta$ is a $P$-dimensional unknown parameter.  We specify a prior $p(\theta)$, and we wish to sample from the posterior $p(\theta \mid X^{(\mathrm{obs})})$. The goal of neural posterior estimation is to provide a tractable approximation to $p(\theta \mid X)$ by leveraging the power of generative neural networks \cite{Biazzo_2022, radev2020bayesflowlearningcomplexstochastic, greenberg2019automaticposteriortransformationlikelihoodfree, Cranmer_2020, papamakarios2019sequentialneurallikelihoodfast, lueckmann2017flexiblestatisticalinferencemechanistic}. In NPE, we approximate the true posterior by fitting a generative neural network to a large set of synthetic training data, consisting of $(\theta, X)$ pairs drawn from the joint distribution $p(\theta) \cdot p(X \mid \theta)$. Thus the task of posterior estimation becomes one of estimating a conditional density using a high-capacity generative model, denoted $q_\phi(\theta \mid X)$, where $\phi$ are the network parameters.  Once trained, the NPE model $q$ serves as a fast amortized inference engine: given the dataset $X^{(\mathrm{obs})}$ actually observed, the network can produce samples from the estimated posterior distribution of $\theta$ by simulating draws from the fitted NPE model, $q_\phi(\theta \mid X^{(\mathrm{obs})})$.  While NPE is especially useful in cases where the likelihood function is intractable or computationally expensive, it remains relevant even in traditional ``white box'' scenarios where we have explicit knowledge of the likelihood function, especially if standard techniques like MCMC are slow or difficult to implement efficiently. 

\paragraph{The simulation phase.}

The simulation phase in NPE consists of repeatedly simulating $\theta^{(m)} \sim p(\theta)$, and then data $X^{(m)}$ conditional on $\theta^{(m)}$, for $m = 1, \ldots, M$. (Typically the number of simulations $M$ is quite large, e.g. $10^6$ or more.)  In most settings, each simulated $X^{(m)}$ is not just a single data point, but rather an entire dataset of (typically vector-valued) observations generated under the assumption that $\theta = \theta^{(m)}$.  We use $N_m$ to refer to the number of samples in simulated dataset $m$, and $D$ to refer to the dimensionality of the sample space, i.e. the support of $p(X \mid \theta)$.  For example, in the common case where individual data points are exchangeable, the dataset $X^{(m)}$ consists of conditionally independent and identically distributed (IID) draws $X^{(m)}_i \in \mathcal{R}^D$ from the data generating process, given the parameter $\theta^{(m)}$:
$$
p(X^{(m)} \mid \theta^{(m)}) = \prod_{i=1}^{N_m} p \left( X^{(m)}_i \mid \theta^{(m)} \right) \, , \quad N_m \sim S \, ,
$$
with obvious modifications for non-IID data scenarios, e.g. sequential data. The tuple $\{ \theta^{(m)}, X^{(m)}\}$ is then a single training instance for the NPE model. Note that the sample size $N_m$ is not necessarily fixed across all simulations but instead varies according to a pre-specified probability distribution $S$. This ensures that the NPE model is exposed to datasets of differing sizes during training, allowing it to learn how the posterior distribution should contract as a function of the sample size, rather than overfit to a single fixed sample size. 

\paragraph{The summary and decoder networks.}

Neural Posterior Estimation (NPE) typically involves a two-stage summary/decoder architecture. The summary and decoder networks work together to approximate the posterior distribution $p(\theta \mid X)$ by first summarizing or encoding the dataset $X$ in terms of a fixed-dimensional representation, and then producing posterior samples based on this representation.

The encoder or summary network $s_\psi(\cdot)$ is a parameterized function that is responsible for mapping the dataset $X^{(m)}$ to a fixed-dimensional vector of $K$ summary statistics.  This summary vector, $s_m \equiv s(X^{(m)})$, can be thought of as a learned analog of a sufficient statistic, although unlike in classical statistical theory, the encoder is trained on simulated data to extract the relevant information about $\theta$; no knowledge of the structural form of $p(X \mid \theta)$ is used.

Formally, let \( \mathcal{X} \) denote the set of all data sets that might be encountered during the simulation phase, defined as a union over all possible sample sizes:
\[
\begin{aligned}
\mathcal{X} &= \bigcup_{N \in \text{supp}(S)} \left\{ \left\{ x_i \right\}_{i=1}^{N} : x_i \sim p(x \mid \theta), \, \theta \in \Theta \right\} \, ,
\end{aligned}
\]
where \( \text{supp}(S) \) represents the support of the distribution \( S \) that determines the (random) sample size for a given data set.  We can then formally define a summary network as a function $s:\mathcal{X} \longrightarrow \mathcal{R}^k$, i.e. a function that encodes any possible data set, regardless of size, as a fixed-dimensional vector that summarizes the data. %Importantly, the encoder must not only summarize the key patterns in the data but also handle the fact that $X^{(m)}$ is a set-valued input whose size may vary from one dataset to the next. The challenge here is to ensure that this variable-sized input can be mapped to a fixed-dimensional vector that is rich enough to encode all relevant information about the posterior for $\theta$, regardless of the sample size $N_m$. 
The structure of the summary network will be problem dependent, and will typically be designed to reflect the type of data encountered (e.g. exchangeable, sequential, spatial). We discuss the choice of summary network in describing our experiments below, considering a range of options proposed across the literature, including DeepSets \cite{bloemreddy2020probabilisticsymmetriesinvariantneural,zaheer2018deepsets,zhang2022setnormequivariantskip}, Janossy pooling \cite{murphy2019janossypoolinglearningdeep}, LSTMs \cite{gersfelix2000learningtoforget}, and set attention \cite{lee2019settransformerframeworkattentionbased}.

Once the encoder has produced a summary statistic $s_m \equiv s_\psi(X^{(m)})$, the decoder network $f(\cdot)$ takes over. The decoder is trained to generate posterior samples for $\theta$, conditional on the summary $s_m$; in other words, it learns the approximate conditional distribution $p(\theta \mid X) \approx q_\phi(\theta \mid s_\psi(X))$. Prior work on NPE decoders has focused on normalizing flows that transform a simple base distribution (e.g. a Gaussian) into the target posterior  via a composition of many simple invertible transformations. This transformation is accomplished by a neural network that learns a flexible mapping between these two distributions, via the relation
$$
\theta \sim q_\phi(\theta \mid X) \iff \theta = f_\phi(z, s_\psi(X))
$$
where $f$ is a learned (deterministic) function parameterized by $\phi$, $z$ is a random draw from the base distribution, and $s_\psi(X)$ is the summary vector by the learned encoder network. By training the network to minimize a divergence between the predicted and true conditional distributions, the decoder learns to approximate the posterior for any input data set $X$.  

There are several other possible methods for neural posterior estimation, each offering different ways to approximate complex posteriors, see e.g. \cite{cranmer2015approximating}, \cite{papamakarios2019sequential} and \cite{uria2016nade}. However, the seminal paper on BayesFlow \cite{radev2020bayesflowlearningcomplexstochastic}  showed that decoders based on normalizing flows consistently outperform other methods, especially in high-dimensional or multimodal problems. This led to their widespread adoption for simulation-based inference.  For this reason, we keep the presentation concise by benchmarking our diffusion-based decoders against normalizing flows, rather than all other NPE methods, which have largely been deprecated in practice by flow-based decoders.

\section{Conditional diffusions for NPE}

In our initial setup, we consider the case where there is only one data sample corresponding to each $\theta$ in order to introduce the functionality of the diffusion decoder. We then extend our method by incorporating a summary network to handle the entire dataset $X$. 

Diffusion models \cite{ho2020denoising,song2020denoising,karras2022elucidating} are powerful generative models that operate by defining a forward diffusion process that gradually transforms the data distribution into a noise distribution. The model is then used to reverse this process, generating novel samples from pure noise.  (In the context of NPE, ``data'' actually refers to the sampled $\theta$ values.) Specifically, we aim to construct a diffusion process $\{\theta_t\}_{t=0}^T$ indexed by the continuous time variable $t \in [0, T]$. This diffusion process should satisfy the conditions $\theta_0 \sim p(\theta \mid X)$ and $\theta_T \sim p_T$, where $p_T$ is typically a Gaussian distribution that is straightforward to sample from. We abbreviate the notation by using $\theta_0$ as $\theta$ when there is no ambiguity. Starting from Gaussian noise, we can then employ a Reverse Ordinary Differential Equation (ODE) to sample from the noise distribution to $\theta_0$, a sample from the target distribution, formulated as:
\begin{align}
    \label{eq:reverse_sde}
    \text{d}\theta = \left[f(\theta,t) - \frac{1}{2}g(t)^2\nabla_\theta\log p_t(\theta \mid X)\right]\text{d}t,
\end{align}
$\nabla_\theta\log p_t(\theta \mid X)\coloneq\nabla_{\theta_t}\log\int p(\theta_t \mid \theta_0)p(\theta_0 \mid X)\text{d}\theta_0$. Here, $p(\theta_t \mid \theta_0)$ is a transition kernel commonly represented as a Gaussian noise addition process, i.e., $\theta_t \sim \mathcal{N}(\alpha(t)\theta_0,\beta(t)^2\bm{I})$, where $\alpha(t)$ and $\beta(t)$ are predefined schedules and correspond to a forward Stochastic Differential Equation (SDE).
\begin{align}
\label{eq:forward_sde}
    \text{d}\theta = f(\theta,t)\text{dt} + g(t)\text{d}w,
\end{align}
where $w$ is Brownian motion. It is proven in \citet{song2020score} that the forward SDE (Eq. \ref{eq:forward_sde}) has the same marginal distribution $p_t(\theta|X)$ as the reverse ODE (Eq. \ref{eq:reverse_sde}) for any gievn $X$. Moreover, $\alpha(t)$ and $\beta(t)$ have a closed-form relationship with $f(\theta,t)$ and $g(t)$. We defer a more detailed discussion to Appendix \ref{sec:diff_training}.

Once the score function $\nabla_\theta\log p_t(\theta \mid X)$ is known for all $t$, we can solve this ODE to sample from the Gaussian noise distribution $\theta_T$ to $\theta_0$.

The score function $\nabla_\theta\log p_t(\theta \mid X)$ can be learned by training a time-dependent mean-prediction model ${\mu_\phi}(\theta_t \mid X,t)$ using the following loss function (Eq~\ref{eq:loss_diffusion}). This relies on the connection between the mean-prediction model and the score function, given by  
\begin{align}
    \nabla_\theta\log p_t(\theta \mid X) \approx -\frac{\theta_t - \alpha(t)\mu_\phi(\theta_t \mid X,t)}{\beta_t^2}.
\end{align}
The corresponding training objective is formulated as:  
\begin{align}
    \gL (\phi)= \E_{p(t,X,\theta_0,\theta_t)} \left[ \lambda(t) \|\mu_\phi(\theta_t \mid X,t) - \theta_0\|_2^2 \right] 
    \label{eq:loss_diffusion}
\end{align}

where $\lambda(t)$ is a positive weighting function, and $p(t)$ is a predefined time schedule used for training. It is proven by \citet{song2020score, karras2022elucidating} that the optimal solution $\mu_\phi(\theta_t \mid X,t)$ of $\gL(\phi)$ will be $\E[\theta_0|\theta_t,X]$.  More details can be found in Appendix \ref{sec:diff_training}. Since $p(\theta_t \mid \theta_0)$ is a Gaussian distribution, this term can be computed analytically. This loss function is used to train the diffusion decoder $\phi$ and serves as a variational bound on the negative log-likelihood as well as the Kullback-Leibler (KL) divergence, as described in Proposition \ref{thm:diffusion_upperbound}. 

\begin{proposition}
\label{thm:diffusion_upperbound}
\begin{align*}
    \E_{p(X)}\left[D_{KL}(p(\theta \mid X)||q_{\phi}(\theta \mid X))\right]\le \gL(\phi) + C
\end{align*}
where $C$ is independent of $\phi$.
\end{proposition}

\begin{proof}[Proof Sketch]
    \begin{align}
        \E_{p(X)}&\left[ D_{KL}(p(\theta \mid X)||q_{\phi}(\theta \mid X))\right] \nonumber \\
        &= \E_{p(X)p(\theta \mid X)}[\log p(\theta \mid X)-\log q_\phi(\theta \mid X)] \nonumber \\
        &= \E_{p(\theta)p(X \mid \theta)}[-\log q_\phi(\theta \mid X)] +C\nonumber \\
        &\leq \gL(\phi)+C\nonumber
    \end{align}
    where the last inequality is proven in \cite{ho2020denoising} and \cite{song2021maximum}. The complete proof can be found in Appendix \ref{sec:diff_proof}.
\end{proof}

This shows that training a diffusion model is equivalent to minimizing an upper bound on the KL divergence between the ground truth posterior $p(\theta \mid X)$ and our decoder $q_{\phi}(\theta \mid X)$. Note from (\ref{eq:loss_diffusion}) that the loss function is nonnegative; therefore, the optimal solution of $\gL(\phi)$, given sufficient data and model capacity, implies that $p(\theta \mid X) = q_{\phi}(\theta \mid X)$ for any $X$.

Next, we consider the case where the decoder must condition on the entire dataset $X$, rather than just a single data point corresponding to $\theta$. Since the size of $X$ may vary, we introduce a parameterized summary network $s_\psi(X)$ as described in Section \ref{sec:preliminaries}. This network computes summary statistics of a fixed dimension, regardless of the number of data points in $X$. We jointly train the summary network and the decoder using the following loss function:
\begin{align}
\label{eq:summary_diffusion}
    \gL(\psi, \phi) &= \E_{p(t,X,\theta_0,\theta_t)}\left[\lambda(t)\|\mu_\phi(\theta_t\mid s_\psi(X),t)- \theta_0\|_2^2\right],
\end{align}
where $s_\psi(X)$ denotes the output of the summary network. Since the summary network must capture sufficient statistics for the dataset input, its architecture is crucial and should be tailored for the structure and probabilistic invariances of the data at hand. Detailed information about the summary network structure is provided in Appendix \ref{sec:summary_network}.

We further establish that jointly training the summary network, parameterized by $\phi$, and the decoder, parameterized by $\psi$, using the diffusion loss $\gL(\psi, \phi)$ remains a valid upper bound on the KL divergence between the true posterior distribution and the estimated posterior distribution, as stated in Proposition \ref{thm:summary_diffusion}.

\begin{proposition}
\label{thm:summary_diffusion}
\begin{align*}
    \E_{p(X)}\left[D_{KL}(p(\theta \mid X) \mid \mid q_{\phi}(\theta\mid s_{\psi}(X)))\right] \leq \gL(\psi, \phi) + C',
\end{align*}
where $C'$ is a constant independent of $\psi$ and $\phi$.
\end{proposition}

A complete proof is deferred to Appendix \ref{sec:summary_diffusion}.

In Algorithm \ref{alg:train_cdiff}, we provide a detailed description of the procedure for training $\psi$ and $\phi$ using the loss function $\gL(\psi, \phi)$, as defined in (\ref{eq:summary_diffusion}). Posterior samples are then generated following the procedure outlined in Algorithm \ref{alg:sample_cdiff}.

\section{Empirical results}
\label{sec:results}

\subsection{Overview of benchmark suite}

Our benchmark problems fall into three categories: no-encoder, IID, and sequential. We provide details of all problems and model settings in Appendix \ref{app:benchmark_problems}; here, we briefly describe the problems and their encoders. 

In the first group (``no-encoder'' problems), we pass either a single data point $X$ directly to the decoder, or a vector of sufficient statistics.  Our goal here is to study pure decoder scenarios which do not require the model to learn an encoder jointly with the decoder.

In the second group (IID problems), each simulated data set $X^{(m)}$ consists of many IID observations from the data generating process, of varying sample sizes $N_m \sim U(N_{\min}, N_{\max})$. In our main experiments, we use an encoder/summary network based on DeepSets \cite{zaheer2018deepsets} for all IID problems. We also ran secondary experiments using attention-based summary networks \cite{lee2019settransformerframeworkattentionbased} and give results in the Appendix \ref{app:transformer}. We attempted to run experiments used summary networks based on Janossy pooling \cite{murphy2019janossypoolinglearningdeep}, but these networks performed very poorly when small enough to be computationally tractable.

In the third group (sequential problems), each data set $X^{(m)}$ consists of sequential data of varying sequence lengths $N_m \sim U(T_{\min}, T_{\max})$. For these problems we use a summary network based on bidirectional LSTMs, as in \cite{radev2020bayesflowlearningcomplexstochastic}, to capture the temporal characteristics of the data. The biLSTM operates as a sequence-to-sequence mapping; to get a fixed-dimensional summary, we concatenate the last biLSTM cell states in both the forward and backward directions.

\begin{table*}[t]
\centering
\begin{small}
\caption{\label{tab:benchmark_results} Performance results across 13 benchmark problems for 10 independent runs: average Wasserstein distance (WD) to U(0,1) across all marginals and runs; Wasserstein distance for worst marginal; and empirical coverage probabilities (ECP) from TARP. Numbers are multiplied by $10^3$. Smaller numbers are better. }
\resizebox{\textwidth}{!}{%
\begin{tabular}{llrrrrrrr}
\toprule
\multirow{2}{*}{Group} & \multirow{2}{*}{Problem} & \multicolumn{3}{c}{cNF} & \multicolumn{3}{c}{cDiff} \\ \cmidrule(lr){3-5} \cmidrule(lr){6-8}
  &  & WD (avg) & WD (worst) & ECP & WD (avg) & WD (worst) & ECP \\ 
\midrule
\multirow{7}{*}{No Encoder} 
  & Sum of cosines & 41.077 & 44.224 & 11.648& 31.922 & 35.453 & 8.861 \\
  & Witch's hat & 103.921 & 123.615 & 25.658 & 26.579 & 41.773 & 7.084 \\
  & Dirichlet multinomial & 65.332 & 74.707 & 8.892 & 26.747 & 39.278 & 6.394\\
  & Socks & 34.813 & 55.560 & 6.647 & 28.114 & 50.397 & 8.873 \\
  & Species sampling & 36.942 & 41.275 & 8.782 & 29.184 & 35.181 & 7.023 \\
  & Poisson gamma & 31.797 & 44.425 & 6.276 & 26.637 & 36.880 & 6.462 \\
\cmidrule(lr){2-8}
& \textbf{average} & 52.314 & 63.301 & 11.317 & \underline{28.197} & \underline{39.160} & \underline{7.450} \\
\midrule
\multirow{4}{*}{IID} 
  & Normal gamma & 29.400 & 37.129 & 8.590 & 26.076 & 31.930 & 8.607 \\
  & Multivariate g-and-k & 54.914 & 84.451 & 10.964 & 29.336 & 53.309 & 10.894 \\
  & Normal wishart & 51.499 & 84.632 & 11.014 & 33.918 & 83.356 & 7.445 \\
  \cmidrule(lr){2-8}
& \textbf{average } & 45.271 & 68.071 & 10.189 & \underline{29.110} & \underline{56.198} & \underline{8.982} \\
\midrule
\multirow{6}{*}{Sequential} 
  & Lotka--Volterra & 163.704 & 177.046 & 39.161 & 59.565 & 159.502 & 10.046 \\
  & fractional BM & 89.143 & 139.194 & 106.446 & 47.211 & 101.710 & 15.814 \\
  & stochastic vol & 26.311 & 56.570 & 15.066 & 27.199 & 59.377 & 19.792 \\
  & Markov switch & 25.219 & 45.245 & 8.987 & 34.386 & 66.200 & 6.953 \\
  & VAR(P) & 31.670 & 55.399 & 16.783 & 25.449 & 51.078 & 49.334 \\
  \cmidrule(lr){2-8}
  & \textbf{average} & 67.209 & 94.491 & 37.289 & \underline{38.762} & \underline{87.973} & \underline{20.388} \\
\bottomrule
\end{tabular}%
}
\end{small}
\end{table*}

\begin{figure*}[h!] % Use [H] to stick the figure exactly where placed
    \centering
    \begin{subfigure}[b]{0.46\columnwidth} % Set subfigure width to fit within a single column
        \includegraphics[width=\textwidth]{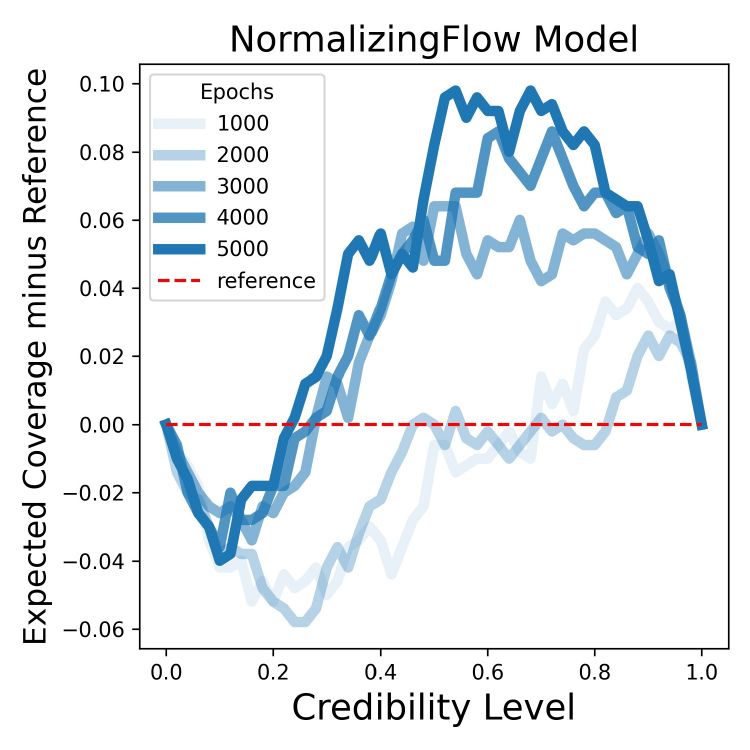}
        \caption{TARP (cNF)}
        \label{fig:figure1}
    \end{subfigure}
    \begin{subfigure}[b]{0.46\columnwidth} % Set subfigure width to fit within a single column
        \includegraphics[width=0.98\textwidth]{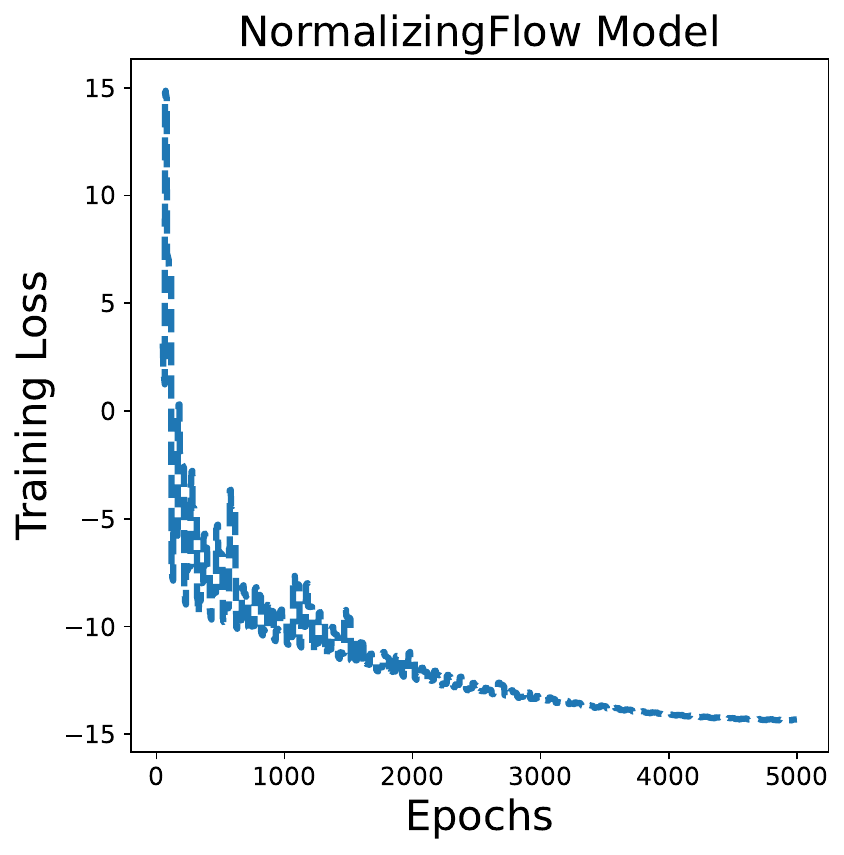}
        \caption{Train loss (cNF)}
    \end{subfigure}
    \begin{subfigure}[b]{0.46\columnwidth} % Set subfigure width to fit within a single column
        \includegraphics[width=\textwidth]{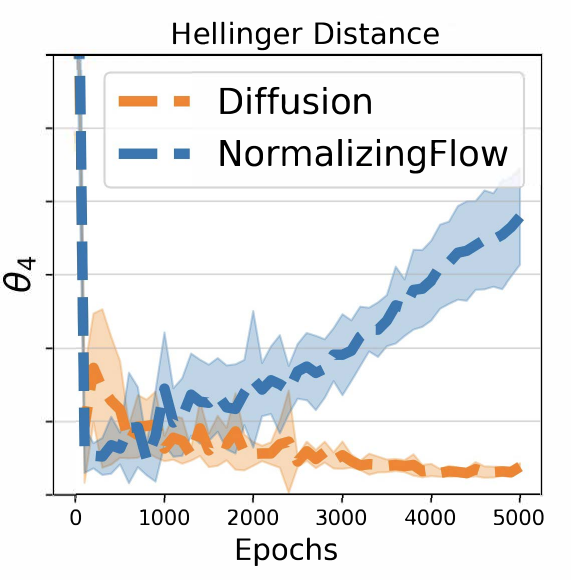}
        \caption{SBC (both)}
    \end{subfigure}
        \begin{subfigure}[b]{0.46\columnwidth} % Set subfigure width to fit within a single column
        \includegraphics[width=\textwidth]{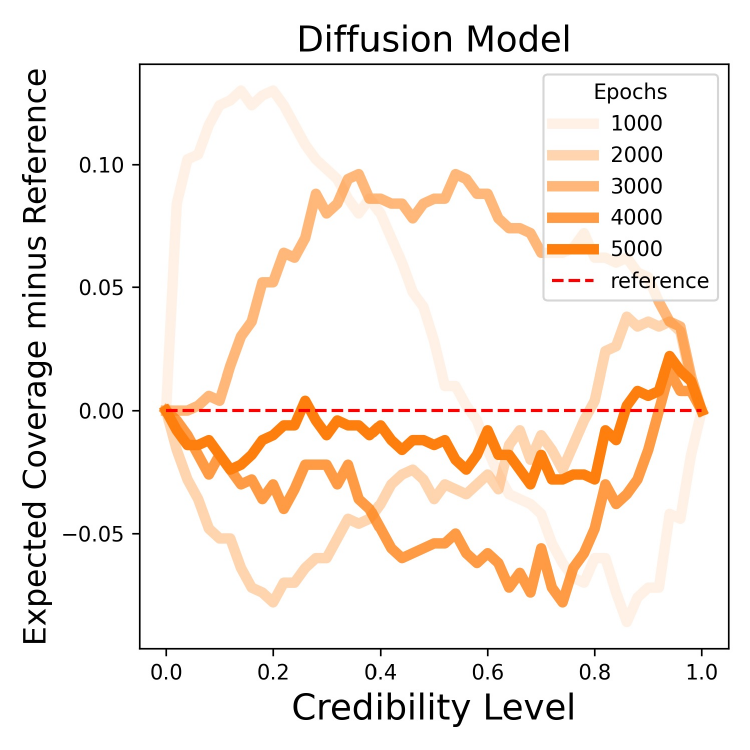}
        \caption{TARP (cDiff)}
        \label{fig:figure2}
    \end{subfigure}
    \caption{Normalizing flows exhibited diverging ECP panel (A) and SBC (panel C) metrics on the fractional Brownian motion problem, despite apparent convergence of the training error (panel B). The difference between expected coverage and the reference coverage should converge to zero across all credibility levels with more training. This difference visibly diverges for normalizing flows (A), but tends toward zero for diffusions (D).  This is confirmed by examining both models' SBC metrics (C) for a single parameter (in this case, the phase of the cosine drift). This metric should also go to 0 with more training.  Qualitatively similar results were observed for the Lotka-Volterra model. More details are deferred to Appendix \ref{app:evaluations}.}
    \label{fig:stability}
\end{figure*}

Conditional diffusions and normalizing flows have very different structures, but we made an effort to construct versions of each model class that had roughly the same number of total parameters (204K for diffusions, 215K for flows), to ensure a fair test. Details of model architectures are in the Appendix \ref{sec:diff_training} and \ref{app:nf}.  We trained each model on the same amount of data, sufficient to ensure convergence of both models. We repeated this 10 times for each problem and averaged the results.   

One operational challenge is that flows and diffusion models optimize different, noncomparable losses.  Moreover, without knowledge of the true posterior, it is impossible to validate NPE methods by calculating test-set loss versus a ground truth.  Instead, we rely on two distinct approaches for measuring the quality of a neural posterior approximation: \emph{simulation-based calibration}, or SBC \cite{talts2020validatingbayesianinferencealgorithms}, which measures the accuracy of each univariate marginal of the posterior approximation; and \emph{tests of accuracy by random points}, or TARP \citep{lemos2023tarp}, which measures the accuracy of the joint posterior approximation.

Both methods give rise to statistics that can compared to a $U(0,1)$ distribution. For SBC, we measure performance using the \emph{average} Wasserstein distance to $U(0,1)$ across all runs and all univariate marginals, as well as the \emph{worst-case} (max) distance over all marginals and all runs.  For TARP, we averaged the distance to $U(0,1)$ across all runs. Details of these metrics are in Appendix \ref{app:validation}.

\subsection{Findings and discussion}
\label{sec:results}
Table \ref{tab:benchmark_results} shows the results of our benchmarking experiments.  Training times were 35\% faster for diffusions (1.752s per batch, on average across all problems) than for normalizing flows (2.674s per batch, on average).  Inference times are slower for diffusions (30.25 milliseconds, vs.~22.47ms, on average), although this is a minor inconvenience; since NPE is an amortized inference method, much more time is devoted to training than to inference. All experiments were conducted on a system with 8 NVIDIA A5000 GPUs, and an AMD EPYC 7702P 64-Core CPU used to simulate data. Each individual training run was done on a single GPU.

The results, along with Figures \ref{fig:cosine_example}--\ref{fig:dirichlet_example}, indicate that conditional diffusions outperform normalizing flows across most problems. In the No Encoder group, conditional diffusions perform better across the board, with lower average Wasserstein distances (WD avg) and worst marginals (WD worst). They particularly excel in challenging problems such as the Witch's Hat, where conditional diffusions achieve a significantly better average and worst Wasserstein distance compared to normalizing flows (26.579 vs. 103.921 for average WD). In the IID group, conditional diffusions also consistently outperform normalizing flows, with lower average WD and ECP. For example, in the Multivariate G-and-K distribution, conditional diffusions achieve a much lower average WD (29.336 vs. 54.914) across the marginals, although the ECP values are similar.  This is potentially explained by the dependence of the ECP metric on the choice of reference distribution; ECP can be notably less sensitive at detecting deviations from \( p(\theta \mid y) \) when the reference distribution is not tuned well. Finally, in the Sequential group, the results are more mixed, although still favorable for diffusions.  While conditional diffusions perform substantially better on Lotka–Volterra and fractional BM, normalizing flows slightly outperform conditional diffusions in the Markov switching model. The diffusion model also exhibited stable convergence on all metrics across all problems, whereas the normalizing flow showed diverging WD and ECP metrics on some problems, indicating poor posterior approximations despite (apparently) convergent training loss. See Figure \ref{fig:stability}.

See Appendix \ref{app:evaluations} for further results. In summary, conditional diffusions generally performed better---often substantially better---across most problems, while normalizing flows performed a bit better in isolated cases. 

\section{Limitations}

Our work has several limitations that merit a brief discussion.  First, current NPE methods typically assume that the ground-truth forward model, i.e., the prior \( p(\theta) \) and the likelihood \( p(X\mid\theta) \), is known exactly. However, in practical applications, the true generative process is often only imperfectly specified, leading to a mismatch between the assumed model used during training and the actual data-generating process. While model misspecification is a fundamental concern in any Bayesian inference procedure, it poses a particularly acute problem for NPE. In classical Bayesian computational methods such as MCMC or variational inference, inference is conducted under the given (possibly misspecified) model, meaning that the resulting posterior distribution is still internally consistent with that model, even if the model itself is incorrect. That is, one at least obtains the ``correct'' Bayesian posterior, given the assumed (but possibly misspecified) likelihood. By contrast, NPE is trained on simulations from the assumed model and learns to generalize to new observations based on that training distribution. If the true data generating process deviates significantly from the assumed forward model, the inference procedure may be forced to make predictions on out-of-distribution data---i.e., queries that fall off the training data manifold. Since deep learning models often fail unpredictably when extrapolating beyond their training distribution, this raises the concern that the inferred posterior may be highly unreliable under model misspecification, potentially leading to severe overconfidence or systematic biases. Investigating the robustness of NPE under such circumstances is beyond the scope of this paper, but remains an important direction for future work. 

Second, although we have demonstrated that jointly training the summary network and the diffusion decoder yields a valid evidence lower bound, the interaction between these two components may be more intricate than our current formulation explicitly captures. In principle, the summary network and decoder could exhibit complex dependencies, where architectural choices or training strategies influence their ability to effectively complement each other. Our current approach optimizes both components jointly under a fixed design, but we have not conducted a comprehensive search over alternative configurations that might better exploit their interaction, i.e. how the decoder leverages the encoded representation. Exploring these potential refinements---both in terms of training procedures and architectural modifications---could lead to more efficient and robust posterior estimation, making this an important avenue for future investigation.

Lastly, we observed several drawbacks in current evaluation methods for simulation-based inference settings. One common evaluation metric is $\mathbb{E}_{p(\theta,X)}[-\log q_\phi(\theta|X)]$, which is feasible for normalizing flows but challenging to compute for diffusion-based decoders. While methods for estimating log-likelihood in diffusion models do exist \cite{song2020score}, they involve integrating along the reverse ODE trajectory, which is computationally expensive, requires discretization, and relies on the Skilling-Hutchinson trace estimator. On the other hand, alternative evaluation methods, such as SBC and TARP, provide diagnostic visual comparisons between the ground-truth and estimated posteriors, but do not offer statistically rigorous tests. Furthermore, existing evaluation techniques typically average performance across all values of $X$, failing to identify specific values of $X$ where the approximated posterior may exhibit poor alignment. Future work could focus on developing more computationally efficient and statistically rigorous evaluation protocols that better measure the fidelity of posterior approximations.

\subsubsection*{Acknowledgements}
We sincerely thank the reviewers for their valuable feedback and insightful discussions, which have greatly contributed to improving our paper.

%\begin{table}[ht]
%\label{table:time}
%\centering
%\caption{Comparison of Model Size, Training Time, and Inference Time for cNF and cDiff.}
%\resizebox{\columnwidth}{!}{%
%\begin{tabular}{lccc}
%\toprule
%Model & Model Size (MB) & Training Time (s per batch) & Inference Time (ms) \\ 
%\midrule
%cNF   & 2.49 & 2.674 & 22.47 \\
%cDiff & 0.093 & 1.752 & 30.25 \\
%\bottomrule
%\end{tabular}%
%}
%\end{table}

\bibliography{references}
\bibliographystyle{abbrvnat}

%%%%%%%%%%%%%%%%%%%%%%%%%%%%%%%%%%%%%%%%%%%%%%%%%%%%%%%%%%%%
\section*{Checklist}
 \begin{enumerate}

 \item For all models and algorithms presented, check if you include:
 \begin{enumerate}
   \item A clear description of the mathematical setting, assumptions, algorithm, and/or model. [Yes/No/Not Applicable].

Yes. The mathematical setting is described throughout the paper. The training and sampling procedures are outlined in Algorithm \ref{alg:train_cdiff} and Algorithm \ref{alg:sample_cdiff}, respectively. Detailed descriptions of the network structure and training strategies can be found in Appendix \ref{sec:diff_training} and Appendix \ref{sec:summary_network}.

\item An analysis of the properties and complexity (time, space, sample size) of any algorithm. [Yes/No/Not Applicable]

Yes. We measure the computation time and model size of our method and baseline models in Section \ref{sec:results}.

   \item (Optional) Anonymized source code, with specification of all dependencies, including external libraries. [Yes/No/Not Applicable]

   No, we have not submitted the source code. However, we will make it publicly available after acceptance.
   
 \end{enumerate}

 \item For any theoretical claim, check if you include:
 \begin{enumerate}
   \item Statements of the full set of assumptions of all theoretical results. [Yes/No/Not Applicable]

   Yes.
   
   \item Complete proofs of all theoretical results. [Yes/No/Not Applicable]

   Yes. The complete proof is provided in Appendix \ref{sec:diff_proof} and Appendix \ref{sec:summary_diffusion}.

   \item Clear explanations of any assumptions. [Yes/No/Not Applicable]    
   
   Yes, the explanations of the assumptions are provided immediately following each assumption statement.
   
 \end{enumerate}

 \item For all figures and tables that present empirical results, check if you include:
 \begin{enumerate}
   \item The code, data, and instructions needed to reproduce the main experimental results (either in the supplemental material or as a URL). [Yes/No/Not Applicable]

    No, we did not submit any code to replicate the figures and main experimental results along with the paper, but we will make it publicly available after acceptance.

   \item All the training details (e.g., data splits, hyperparameters, how they were chosen). [Yes/No/Not Applicable]

    Yes, all training details are thoroughly described in Appendix \ref{sec:diff_training} and Appendix \ref{sec:summary_network}.

    \item A clear definition of the specific measure or statistics and error bars (e.g., with respect to the random seed after running experiments multiple times). [Yes/No/Not Applicable]

    Yes, we have run our experiments 10 times with different random seeds, and the results reported in Table \ref{tab:benchmark_results} are the means of all runs.
    
    \item A description of the computing infrastructure used. (e.g., type of GPUs, internal cluster, or cloud provider). [Yes/No/Not Applicable]

    Yes. The equipment details are provided in Section \ref{sec:results}.

 \end{enumerate}

 \item If you are using existing assets (e.g., code, data, models) or curating/releasing new assets, check if you include:
 \begin{enumerate}
 
   \item Citations of the creator If your work uses existing assets. [Yes/No/Not Applicable]

    Not Applicable. We don't use any assets.
   
   \item The license information of the assets, if applicable. [Yes/No/Not Applicable]

Not Applicable. We don't use any assets.

   \item New assets either in the supplemental material or as a URL, if applicable. [Yes/No/Not Applicable]

No. We do not provide access to our new assets at this stage. The assets will be made publicly available once the paper is accepted for publication.
   
   \item Information about consent from data providers/curators. [Yes/No/Not Applicable]

Not Applicable. No data provided outside.
   
   \item Discussion of sensible content if applicable, e.g., personally identifiable information or offensive content. [Yes/No/Not Applicable]

Not Applicable.
   
 \end{enumerate}

 \item If you used crowdsourcing or conducted research with human subjects, check if you include:
 \begin{enumerate}
   \item The full text of instructions given to participants and screenshots. [Yes/No/Not Applicable]

Not Applicable.
   
   \item Descriptions of potential participant risks, with links to Institutional Review Board (IRB) approvals if applicable. [Yes/No/Not Applicable]

Not Applicable.

   \item The estimated hourly wage paid to participants and the total amount spent on participant compensation. [Yes/No/Not Applicable]

Not Applicable.

 \end{enumerate}

 \end{enumerate}

\input{supplement}
\end{document}

%% file: math_commands.tex
\usepackage{amsmath,amsfonts,bm}

\def\eqref#1{equation~\ref{#1}}
\def\1{\bm{1}}

\DeclareMathAlphabet{\mathsfit}{\encodingdefault}{\sfdefault}{m}{sl}
\SetMathAlphabet{\mathsfit}{bold}{\encodingdefault}{\sfdefault}{bx}{n}

\def\gL{{\mathcal{L}}}

\def\gN{{\mathcal{N}}}

\newcommand{\E}{\mathbb{E}}

%% file: supplement.tex
\onecolumn
\appendix
\begin{center}
    \Large{\textbf{Conditional diffusions for neural posterior estimation: \\
Supplementary Materials}}
\end{center}

\section{Training and inference algorithms for conditional diffusions}

The main training loop for conditional diffusions, wherein we jointly train a summary network and diffusion decoder, is described in Algorithm \ref{alg:train_cdiff}.

\begin{algorithm}[H]
\caption{\label{alg:train_cdiff}cDiff Training}
\begin{algorithmic}[1]
\State \textbf{Input:} $p(\theta)$, $p(X \mid \theta)$, $\lambda(t)$, $p(\theta(t) \mid \theta(0))$
\State \textbf{Output:} summary network $s_\psi$, diffusion network $\mu_\phi$
\State Initialize networks $s_\psi$ and $\mu_\phi$
\While{not converged}
    \State Sample $\theta\sim p(\theta)$
    \State Sample number of observations $N \sim \text{Unif}[N_{\text{min}},N_{\text{max}}] $
    \State Sample $X=\{X_i\}_{i=1}^N\sim p(X \mid \theta)$
    \State Calculate summary $s = s_{\psi}(X)$
    \State Calculate loss $\gL(\psi,\phi)$, given $(\theta, s)$ (Eq \ref{eq:summary_diffusion})
    \State Take gradient descent step on $\psi,\phi$
\EndWhile
\State Output $s_\psi$ and $\mu_\phi$
\end{algorithmic}
\end{algorithm}

After training the summary network $s_\psi$ and $ \mu_\phi$, we are ready to sample from $p(\theta \mid X)$ for a given dataset $X$, as described in Algorithm \ref{alg:sample_cdiff}. Note that the sampling process applies to any general diffusion process, such as VP, VE \cite{song2020score}, or EDM \cite{karras2022elucidating}. In our paper, we utilize the EDM \cite{karras2022elucidating} training and sampling process, which is detailed in Appendix \ref{sec:diff_training}, where, for notational simplicity, we define $\theta_t \coloneqq \{\theta^{[l]}_t\}_{l=1}^L$.

\begin{algorithm}
\caption{\label{alg:sample_cdiff}cDiff Sampling}
\begin{algorithmic}[1]
\State \textbf{Input:} $s_\psi$,$\mu_\phi,X$, number of posterior sample $L$, EDM schedule $\{\sigma_t\}_{t=0}^T$
\State \textbf{Output:} $\{\theta^{[l]}\}_{l=1}^L$ 
\State $\theta_T\coloneqq \{\theta^{[l]}_T\sim \gN(0,\bm I)\}_{l=1}^L$
\State $s\leftarrow s_\psi(X)$
\For{$t\in \{T,\dots,1\}$}
    \State $\hat\theta_0\leftarrow \mu_\phi(\theta_t\mid s,t)$
    \State $\theta_{t-1}\leftarrow\theta_t - \frac{\sigma_t-\sigma_{t-1}}{\sigma_t}(\theta_t-\hat\theta_0)$
\EndFor
\State Return $\theta_0$
\end{algorithmic}
\end{algorithm}

\clearpage

\section{Diffusion model propositions and implementation details}

\subsection{Proof of Proposition \ref{thm:diffusion_upperbound}}
\label{sec:diff_proof}

\subsubsection{Preliminaries} Here, we first illustrate the connection between score function prediction, mean prediction, and noise prediction. For notational simplicity, we use $\mu(\theta_t,t)$ for the mean prediction, and $\epsilon(\theta_t,t)$ for the noise prediction. By considering the process of adding noise:
\begin{align*}
    \theta_t = \alpha(t)\theta_0 + \beta(t)\epsilon \qquad \epsilon \sim \gN(0,\bm{I}),
\end{align*}
we can train these functions using the following loss:
\begin{align*}
&\E_{p(t),p(\theta),p(X \mid \theta)} \E_{p(\epsilon)}\left[\lambda(t) \|{\epsilon_\phi}(\theta_t \mid X,t)-\epsilon\|_2^2\right] \\
%&\E_{p(t),p(\theta),p(X \mid \theta)} \E_{p(\theta_t \mid \theta_0)}\left[\lambda'(t) \|{z_\phi}(\theta_t \mid X,t)-\nabla_{\theta_t}\log p(\theta_t \mid \theta_0)\|_2^2\right] \\
&\E_{p(t),p(\theta),p(X \mid \theta)} \E_{p(\theta_t \mid \theta_0)}\left[\lambda''(t) \|{\mu_\phi}(\theta_t \mid X,t)-\theta_0\|_2^2\right] \\
\end{align*}
It is proven by \citet{song2020denoising, song2020score, ho2020denoising, ho2022classifier} that the optimal solution of these loss functions has the following connection:
\begin{align*}
    \mu(\theta_t,t) = \frac{\theta_t - \beta(t)\cdot \epsilon(\theta_t,t)}{\alpha(t)}.
\end{align*}
Since the optimal solution corresponds to different reparameterizations of the mean-prediction function $\mu_\phi(\theta_t,t)$, we can use any of these reparameterizations to illustrate the proof.

\subsubsection{Complete Proof of Proposition \ref{thm:diffusion_upperbound}}

\begin{lemma}
\label{lemma:elbo}
    \begin{align*}
        \E_{p(X)} \E_{p(\theta \mid X)}[-\log p_\phi(\theta \mid X)] \le \sum_{t=1}^T \E_{p(X)} \E_{p(\theta \mid X)p(\epsilon)}[\lambda(t)\|\epsilon_\phi(\theta_t \mid X,t)-\epsilon\|_2^2] + C.
    \end{align*}
\end{lemma}

\begin{proof}
    By Equation (3, 8) in \citet{ho2020denoising}, we can adopt the following inequality: For any given $X$, the following holds:
    \begin{align*}
        \E_{p(\theta \mid X)}[-\log p_\phi(\theta \mid X)] \le \sum_{t=1}^T \E_{p(\theta \mid X)p(\epsilon)}[\lambda(t)\|\epsilon_\phi(\theta_t \mid X,t)-\epsilon\|_2^2] + C.
    \end{align*}
    Thus, it is straightforward to conclude that:
    \begin{align*}
        \E_{p(X)} \E_{p(\theta \mid X)}[-\log p_\phi(\theta \mid X)] \le \sum_{t=1}^T \E_{p(X)} \E_{p(\theta \mid X)p(\epsilon)}[\lambda(t)\|\epsilon_\phi(\theta_t \mid X,t)-\epsilon\|_2^2] + C.
    \end{align*}
\end{proof}

\begin{proof}[Complete Proof of Proposition \ref{thm:diffusion_upperbound}]
    \begin{align*}
        \E_{p(X)}&\left[ D_{KL}(p(\theta \mid X) \| q_{\phi}(\theta \mid X))\right]  \\
        &= \E_{p(X)p(\theta \mid X)}[\log p(\theta \mid X) - \log p_\phi(\theta \mid X)]  \\
        &= \E_{p(X)p(\theta \mid X)}[-\log p_\phi(\theta \mid X)] + C \\
        &\leq  \sum_{t=1}^T \E_{p(X)p(\theta \mid X)p(\epsilon)}[\lambda(t)\|\epsilon_\phi(\theta_t \mid X,t)-\epsilon\|_2^2] + C' \\
        &=  \sum_{t=1}^T \E_{p(\theta)p(X \mid \theta)p(\epsilon)}[\lambda(t)\|\epsilon_\phi(\theta_t \mid X,t)-\epsilon\|_2^2] + C',
    \end{align*}
    where the last inequality follows from Lemma \ref{lemma:elbo}. However, this inequality is not efficient for training, as it requires summing over all time steps for every condition $X$. Then, following Equations (12-13) or Equations (15-16) of \citet{kingma2021variational}, when $\lambda(t) = \text{SNR}(t-1)-\text{SNR}(t)$ with $\text{SNR}(t) = \frac{\alpha(t)}{\beta(t)}$, we have:
    \begin{align*}
        \sum_{t=1}^T \E_{p(\theta)p(X \mid \theta)p(\epsilon)}&[\lambda(t)\|\epsilon_\phi(\theta_t \mid X,t)-\epsilon\|_2^2] \\
        &= \frac{T}{2}\E_{p(X),p(\theta \mid X)p(\epsilon),t \sim U\{1,T\}}[\lambda(t)\|\epsilon_\phi(\theta_t \mid X,t)-\epsilon\|_2^2] = \frac{T}{2}\gL(\phi).
    \end{align*}
    Thus, we can conclude that:
    \begin{align*}
        \E_{p(X)}&\left[ D_{KL}(p(\theta \mid X) \| q_{\phi}(\theta \mid X))\right] \le \frac{T}{2}\gL(\phi) + C'.
    \end{align*}

Since $T$ is a known constant, we can absorb the factor $\frac{T}{2}$ into $\gL(\phi)$ without affecting the optimization process. Consequently, Proposition \ref{thm:diffusion_upperbound} holds.

\end{proof}

\subsection{Diffusion Training and Inference Details}
\label{sec:diff_training}

We illustrate the training and inference details for completeness regarding our use of the EDM diffusion framework. The complete details and justification can be referred to in \citet{karras2022elucidating}.

\subsubsection{Training Details}

\paragraph{Noise Schedule}
We illustrate the EDM diffusion training schedule in our setting. First, we need to define some prespecified parameters: $\sigma_{\text{data}} = 0.5$, $\sigma_{\text{min}} = 0.002$, $\sigma_{\text{max}} = 80$. The noise schedule is defined by $\theta_t = \alpha(t)\theta_0 + \beta(t) \epsilon$, where $\epsilon \sim N(0, \bm{I})$. We set $\alpha_t = 1$ and $\beta(t) = t$. $\log(t)$ is sampled from $\mathcal{N}(-1.2, 1.2^2)$.

\paragraph{Denoiser} We are training the diffusion model using the mean prediction loss; thus, the denoiser $\mu_\phi$ is defined as follows. For notational simplicity, we set $\beta(t) = \sigma$. The denoiser $\mu_\phi$ is then defined as:
$$
\mu_\phi(\theta_t|t,X) = c_{\text{skip}}(\sigma) \theta_t + c_{\text{out}}(\sigma) F_\phi(c_{\text{in}}(\sigma) \theta_t, c_{\text{noise}}(\sigma) | X),
$$
where $\sigma = \sigma_t = t$ and $F_\phi$ represents the raw neural network layer. We also define:
\begin{align*}
    c_{\text{skip}}(\sigma) &= \frac{\sigma_{\text{data}}^2}{\sigma^2 + \sigma_{\text{data}}^2}, ~~
    c_{\text{out}}(\sigma) = \frac{\sigma \cdot \sigma_{\text{data}}}{\sqrt{\sigma^2 + \sigma_{\text{data}}^2}}, \\
    c_{\text{in}}(\sigma) &= \frac{1}{\sigma^2 + \sigma_{\text{data}}^2}, ~~
    c_{\text{noise}}(\sigma) = \frac{1}{4} \log(\sigma).
\end{align*}

For $F_\phi$, we use a 4-layer MLP with Positional Embedding for the input $c_{\text{noise}(\sigma)}$.

\paragraph{Weight Schedule}
The final loss is given by:
\begin{align*}
    \mathbb{E}_{\sigma, \theta, X, \epsilon} \left[\lambda(\sigma) c_{\text{out}}^2(\sigma) \left\| F_\phi\left[c_{\text{in}}(\sigma) \cdot (\theta + {\epsilon}), c_{\text{noise}}(\sigma) | X\right] - \frac{1}{c_{\text{out}}(\sigma)} \left[ \theta_t - c_{\text{skip}}(\sigma) \cdot (\theta_t + {\epsilon}) \right] \right\|_2^2 \right],
\end{align*}
where $\lambda(\sigma) = \frac{1}{c_{\text{out}}^2(\sigma)}$.

\subsubsection{Inference Details}

By Equation (1) in \citet{karras2022elucidating}, the EDM diffusion process corresponds to an ODE:
\begin{align*}
    \text{d}\theta = -\dot\sigma(t)\sigma(t)\nabla_\theta\log p(\theta;\sigma(t),X)\text{d}t,
\end{align*}
where the score function $\nabla_\theta\log p(\theta;\sigma(t),X)$ can be obtained by $\left(\mu(\theta_t;\sigma(t),X) - \theta_t\right)/\sigma^2$. We use an Euler ODE solver, and the time step discretization is set by:

\begin{align*}
    \sigma_{\max}^{1/\rho} + \frac{i}{N-1}(\sigma_{\min}^{1/\rho} - \sigma_{\max}^{1/\rho})^\rho,
\end{align*}
where $\rho = 7$, $N = 18$, and $\sigma(t) = t$.

\newpage

\section{Summary Network}

In this section, we introduce how the training of the summary network is combined with the training of the decoder, and how we design the summary network.

\subsection{Proof of Proposition \ref{thm:summary_diffusion}}
\label{sec:summary_diffusion}

The proof follows the logic and notation of the standard diffusion process in \cite{ho2020denoising} and \cite{luo2022understanding}, generalized to the conditional diffusion setting with a trainable summary network $\psi$. We denote the forward diffusion process as $q(\theta_T|\theta_0)$, where $\psi$ represents the summary network and $\phi$ the diffusion encoder. Our proof follows the denoising diffusion probabilistic model (DDPM) framework but can be easily generalized to other diffusion schedules.

\begin{equation}
q(\theta_t | \theta_{t-1}) = \mathcal{N}(\theta_t; \sqrt{\alpha_t}\theta_{t-1}, (1 - \alpha_t)\mathbf{I}),
\end{equation}

where $\bar\alpha_t = \prod_{i=1}^t\alpha_i$, and $\{\alpha_i\}_{i=1}^T$ is a predefined sequence.

Following Eq(5) in \cite{ho2020denoising}, the evidence lower bound (ELBO) is expressed as:

\begin{align*}
\E_{p(\theta,X)}[-\log q_\phi(\theta|s_\psi(X))] \le &
D_{KL}\big(q(\theta_T|\theta_0) \| q_\phi(\theta_T|s_\psi(X))\big) - \mathbb{E}_{q(\theta_1|\theta_0)} \big[ \log q_\phi(\theta_0|\theta_1,s_\psi(X)) \big] \\
& + \sum_{t=2}^{T} \mathbb{E}_{q(\theta_t|\theta_0)} \big[ D_{KL}\big(q(\theta_{t-1}|\theta_t, \theta_0) \| q_\phi(\theta_{t-1}|\theta_t,s_\psi(X))\big) \big].
\end{align*}

The first term, $D_{KL}\big(q(\theta_T|\theta_0) \| q_\phi(\theta_T|s_\psi(X))\big)$, is zero under the standard diffusion assumption. Adding sufficient Gaussian noise in the forward process ensures that $q(\theta_T|\theta_0)$ approximates a standard Gaussian distribution. Likewise, $q_\phi(\theta_T|s_\psi(X))$ follows a standard Gaussian distribution at initialization, making this term vanish (see Eq. 32–33 in \cite{luo2022understanding}).

The second term simplifies as follows:

\begin{align*}
-\mathbb{E}_{q(\theta_1|\theta_0)} \big[ \log q_\phi(\theta_0|\theta_1,s_\psi(X)) \big] 
&=  \mathbb{E}_{q(\theta_1|\theta_0)} \big[ w_1 \|\theta_0 - \mu_\phi(\theta_1| s_\psi(X),t=1)\|_2^2 \big] +C_1,
\end{align*}

where $w_1$ is a weight determined by the predefined diffusion process and does not involve any trainable parameters, and $C_1$ is a constant independent of trainable parameters.

For the third term, following Eq. 87-92 in \cite{luo2022understanding}, $D_{KL}\big(q(\theta_{t-1}|\theta_t, \theta_0) \| p_\phi(\theta_{t-1}|\theta_t,s_\psi(X))\big)$, we denote the predicted posterior mean as $\mu_{\phi,\psi}(\theta_t, t)$, shorthand for $\mu_\phi(\theta_t|s_\psi(X),t)$.

\begin{align*}
D_{KL}&\big(q(\theta_{t-1} | \theta_t, \theta_0) \| q_\phi(\theta_{t-1} | \theta_t,s_\psi(X))\big) \\
&=  D_{KL}\big(\mathcal{N}(\theta_{t-1}; \mu_q, \Sigma_q(t)) \| \mathcal{N}(\theta_{t-1}; \mu_{\phi,\psi}, \Sigma_q(t))\big) \\
&= \frac{1}{2 \sigma_q^2(t)} \bigg\| \frac{\sqrt{{\alpha}_t}(1 - \bar{\alpha}_{t-1})\theta_t 
+ \sqrt{\bar{\alpha}_{t-1}}(1 - \alpha_t)\mu_{\phi,\psi}(\theta_t, t)}{{1 - \bar{\alpha}_t}} - \frac{\sqrt{{\alpha}_t}(1 - \bar{\alpha}_{t-1})\theta_t + \sqrt{\bar{\alpha}_{t-1}}(1 - \alpha_t)\theta_0}{{1 - \bar{\alpha}_t}} \bigg\|_2^2  +  C_2\\
&=  \frac{1}{2 \sigma_q^2(t)} \frac{\bar{\alpha}_{t-1}(1 - \alpha_t)^2}{(1 - \bar{\alpha}_t)^2} 
\bigg\| \mu_{\phi,\psi}(\theta_t, t) - \theta_0 \bigg\|_2^2 +C_2\\
&=  w_t
\bigg\| \mu_{\phi,\psi}(\theta_t, t) - \theta_0 \bigg\|_2^2+C_2
\end{align*}

where $\sigma_q^2(t) = \frac{(1 - \alpha_t)(1 - \bar{\alpha}_{t-1})}{1 - \bar{\alpha}_t}$ and $\Sigma_q(t) = \sigma_q^2(t)\mathbf{I}$. The weight $w_t$ depends on the predefined noise scheduling sequence $\{\alpha_i\}_{i=1}^T$.

Combining the analyses of the three terms, we obtain:

\begin{align*}
\E_{p(\theta,X)}[-\log q_\phi(\theta|s_\psi(X))] &\le \sum_{t=1}^T w_t \bigg\| \mu_{\phi,\psi}(\theta_t, t) - \theta_0 \bigg\|_2^2 +C_1+C_2.
\end{align*}

Thus, as standard diffusion training proceeds, $\phi$ and $\psi$ can be optimized by minimizing the following loss:

\begin{align*}
\mathbb{E}_{t,\theta,X}\mathbb{E}_{q(\theta_t|\theta_0)} \left[ w_t \bigg\| \mu_{\phi}(\theta_t|s_\psi(X),t) - \theta_0 \bigg\|_2^2 \right].
\end{align*}

Finally, we obtain:

\begin{align*}
    \mathbb{E}_{p(X)}\left[D_{KL}(p(\theta \mid X) \| q_{\phi}(\theta \mid s_{\psi}(X)))\right] 
    &=
    \mathbb{E}_{p(\theta)p(X|\theta)}\left[-\log q_{\phi}(\theta \mid s_{\psi}(X))\right]+C_3\\
    &\le \mathbb{E}_{t,\theta,X}\mathbb{E}_{q(\theta_t|\theta_0)}\left[w_t\| \mu_{\phi}(\theta_t|s_\psi(X),t) - \theta_0 \|_2^2\right]+C' = \mathcal{L}(\psi,\phi)+C'.
\end{align*}

where $C'\coloneqq C_1+C_2+C_3$. This loss function is equivalent to that presented in Eq.~\ref{eq:loss_diffusion} of the paper.

\subsection{Summary Network Design}
\label{sec:summary_network}

We use two different structures for the summary network to handle different data types. We refer to the summary network for IID data as DeepSet Summary, and for time series (sequential) data, we use the BiLSTM summary.

\subsubsection{Network Structure for DeepSet Summary}

The main principle of the DeepSet Summary network is based on \cite{bloemreddy2020probabilisticsymmetriesinvariantneural} which utilized the following transformation:

\begin{align*}
    s_\psi(X) = s_{\psi_1}\left(\sum_{i=1}^N s_{\psi_2}(X_i)\right),
\end{align*}

where both $s_{\psi_1}$ and $s_{\psi_2}$ are parameterized as neural networks. This structure ensures that the summary statistics are exchangeable for the data and can capture a wide range of usual sufficient statistics, such as the mean and standard deviation. Additionally, to promote training stability, we normalize $X$ by its sample mean and standard deviation. The normalized data, mean, standard deviation, and $N$ are then input into $s_{\psi_1}$. The learnable scale parameters applied to $N$, the mean, and the standard deviation further contribute to stable training.

\subsubsection{Network Structure for BiLSTM Summary}

A BasicConv1D layer is first applied, performing a one-dimensional convolution that transforms the input data into a higher-dimensional space using a kernel size of 1. The higher-dimensional data is then fed into a BiLSTM network, and the last hidden state is extracted and used as the summary information for the sequential dataset.

\subsection{Discussion about Set Transformer}
\label{app:transformer}

One promising direction to improve the summary network is to use the Transformer architecture \cite{vaswani2017attention} to compute summary statistics, as in \citet{lee2019set}. One advantage of the Transformer-based summary network is its ability to retain long-term information without the forgetting phenomenon observed in LSTMs when dealing with long time-series sequences. Additionally, it can handle both IID and sequential data, eliminating the need to design different summary networks for different data types. 

We implemented Set Transformer as a potential summary network and conducted some preliminary investigations, with the results shown in Table \ref{tab:transformer}. The results indicate that the Set Transformer improves performance on some metrics but not all. Exploring how to integrate the Set Transformer summary network with other decoders will be an interesting direction for future work.

\begin{table*}[ht]
\centering
\caption{\label{tab:transformer}Performance Results Across 13 Problems, Deep Set encoder vs Transformer encoder: per-margin average Wasserstein, per-margin worst Wasserstein, and ECP.}
\resizebox{\textwidth}{!}{%
\begin{tabular}{ccccccccc}
\toprule
\multirow{2}{*}{Group} & \multirow{2}{*}{Problem} & \multicolumn{3}{c}{cNF} & \multicolumn{3}{c}{cDiff} \\ \cmidrule(lr){3-5} \cmidrule(lr){6-8}
  &  & WD (avg) & WD (worst) & ECP & WD (avg) & WD (worst) & ECP \\ 
\midrule
\multirow{8}{*}{IID} 
  & normal gamma & 29.400 & 37.129 & 8.590 & 26.076 & 31.930 & 8.607 \\
  & g and k & 54.914 & 84.451 & 10.964 & 29.336 & 53.309 & 10.894 \\
  & normal wishart & 51.499 & 84.632 & 11.014 & 33.918 & 83.356 & 7.445 \\
  \cmidrule(lr){2-8}
& \textbf{average (Deepset)} & 45.271 & 68.071 & 10.189 & \underline{29.110} & \underline{56.198} & \underline{8.982} \\
\cmidrule(lr){2-8}
  & normal gamma & 28.462 & 32.618 & 17.320 & 23.981 & 24.730 & 17.880 \\  
& g and k & 77.732 & 82.485 & 10.520 & 23.774 & 37.616 & 9.120 \\
& normal wishart & 46.395 & 72.077 & 28.033 & 35.602 & 56.012 & 156.740 \\
\cmidrule(lr){2-8}
& \textbf{average (SetTransformer)} & 50.196 & 62.393 & 18.618 & \underline{27.119} & \underline{39.453} & 61.247\\
\bottomrule
\end{tabular}%
}
\end{table*}

\section{Normalizing Flow structure}
\label{app:nf}

We use a conditional normalizing flow with affine coupling layers as described in \citet{papamakarios2021normalizing}, with a summary dimension of 256, 32 flows, and $\alpha=0.1$, where $\alpha$ represents the soft clamping of the scales in the affine transformations.

\newpage
\section{Validating neural posterior estimates}
\label{app:validation}

Our first approach to validation is simulation-based calibration, or SBC \cite{talts2020validatingbayesianinferencealgorithms}.  SBC uses the fundamental fact that the "expected posterior is the prior," i.e.~$p(\theta) = \int_{\mathcal{X}} p(\theta \mid X) \ p(X) \ dX$, where $p(X) = \int p(X \mid \theta) p(\theta) d \theta$ is the marginal distribution of the data encountered during the simulation phase. In SBC, we check whether this relation also holds for the approximate posterior $q_\phi(\theta \mid X)$. To do so, we draw calibration samples as follows.  For $c = 1, \ldots, C$:
\begin{compactenum}
    \item Sample $\theta^{\star} \sim p(\theta)$, $X \sim p(X \mid \theta^{\star})$
    \item Sample $\tilde{\theta}^{[1]}, \ldots \tilde{\theta}^{[L]} \sim q_\phi(\theta \mid X)$.
    \item For each margin $j$ of $\theta$, calculate $U_j = L^{-1} \sum_{l=1}^L\mathbf{1}(\theta_j^{\star} \leq \tilde{\theta}^{[l]}_j)$.  
\end{compactenum}
If $q_\phi(\theta \mid X) = p(\theta \mid X)$, then each $U_j$ will be uniformly distributed on the unit interval \cite{talts2020validatingbayesianinferencealgorithms}. We construct the empirical distribution of $U$ for each univariate marginal and we calculate Wasserstein distance, total-variation (TV) distance, and Hellinger distance to a $U(0,1)$ distribution. We measure performance using the \emph{average} distance across all runs and all univariate marginals, as well as the \emph{worst-case} (maximum) distance over all marginals and all runs.

The downside of SBC is that it works only for univariate marginals. Our second validation metric is \emph{tests of accuracy by random points}, or TARP \citep{lemos2023tarp}, which measures the accuracy of the joint posterior.  TARP is based on a necessary and sufficient condition for checking the accuracy of a posterior approximation via empirical coverage probabilities.  In TARP, for each simulated pair $(\theta^\star, X)$, we sample a reference point $\theta_r$ from a chosen reference distribution $p_r(\theta)$. Then we sample many $\tilde{\theta}_l$ values from the approximate posterior $q_{\phi}(\theta \mid X)$, given the sampled $X$ value. We then calculate the Euclidean distance $d(\tilde{\theta}_l, \theta_r)$ and compare it to $d(\tilde{\theta}_l, \theta^\star)$. Define $U$ as the fraction of $\tilde{\theta}_l$ values that fall closer to $\theta_r$ than to $\theta^\star$; from \cite{lemos2023tarp}, $U$ should be $U(0,1)$ under the true posterior. As with SBC, we measure the distance of $U$'s empirical distribution to its theoretical uniform distribution.  

\newpage

\section{Details of benchmark problems}
\label{app:benchmark_problems}

\subsection{Summary}

Here, we list all the benchmark simulation processes we used and briefly introduce their features. Then, we illustrate the simulation details for each problem in their individual sections, including hyperparameters and preprocessing. The main goal of preprocessing is to ensure that the inputs $y$ and $\theta$ to the neural networks are unconstrained or basically within the range $[-1,1]$.

\begin{table*}[ht]
    \centering
    \begin{tabular}{|l|l|p{7cm}|}
        \hline
        \textbf{Category} & \textbf{Problem} & \textbf{Key Features} \\
        \hline
        \multirow{6}{*}{No-encoder problems} 
        & Sum of cosines & Multimodality, non-identifiability \\
        & Witch's hat & High entropy, sharp transitions \\
        & Dirichlet-multinomial & Closed/bounded parameter space  \\
        & Poisson-gamma & Conjugate \\
        & Socks & Weak identification \\
        & Species sampling & Weak identification \\
        \hline
        \multirow{3}{*}{IID problems} 
        & Normal-gamma & Conjugate \\
        & Normal-Wishart & Conjugate \\
        & Multivariate G-and-K & Heavy tails, strong upper-tail dependence \\
        \hline
        \multirow{5}{*}{Sequential problems} 
        & Lotka-Volterra & Weak identifiability \\
        & fBM with cosine drift & Long-run dependence, multimodality \\
        & Stochastic volatility & Long-run dependence \\
        & Markov-switching factor model & Sharp changes, bounded support \\
        & VAR(P) & Common econometrics model \\
        \hline
    \end{tabular}
    \caption{\label{tab:benchmark_problems}Summary of benchmarking problems and their key features. Full descriptions of each problem are provided in the appendix.}
\end{table*}

\subsection{Sum of Cosines}

\paragraph{Sampling Process} The sampling model uses a prior distribution \( p(\theta) \) and a likelihood \( p(X \mid \theta) \). The prior distribution is defined over the parameters \(\theta_1\) and \(\theta_2\) as follows:

\[
\theta_1 \sim \text{Uniform}(-1, 1), \quad \theta_2 \sim \text{Uniform}(-1, 1).
\]

Given a sample \(\theta = (\theta_1, \theta_2)\), the likelihood \( p(X \mid \theta) \) is defined by the following data generation process:

\[
\mu = \cos(\theta_1 \pi - \theta_2 \pi) + \cos(2 \theta_1 \pi + \theta_2 \pi) + \cos(3 \theta_1 \pi - 4 \theta_2 \pi),
\]
\[
X \sim \mathcal{N}\left(\mu, 1\right).
\]

\paragraph{Preprocessing} The observed data \( X \) is then scaled by a factor of \(\text{scale}\), resulting in:

\[
X_{\text{scaled}} = \frac{X}{\texttt{scale}}.
\]

\paragraph{Hyperparameters} \(\texttt{scale} = 4\)

\subsection{Witch’s hat}

\paragraph{Sampling Process} The Witch's Hat distribution is a mixture model combining a uniform distribution and a multivariate Gaussian distribution. Given a set of parameters \(\theta \in [0.1, 0.9]^d\), the distribution is defined as:

\[
p(X \mid \theta) = \delta \cdot \text{Uniform}(X; 0, 1) + (1 - \delta) \cdot \mathcal{N}(X; \theta, \sigma^2 \bm I_d),
\]

where \( \text{Uniform}(X; 0, 1) \) denotes a uniform distribution over the \(d\)-dimensional unit hypercube, and \(\mathcal{N}(X; \theta, \sigma^2 \bm I_d)\) represents a multivariate Gaussian distribution centered at \(\theta\) with covariance matrix \(\sigma^2 \bm I_d\).

The prior distribution for \(\theta\) is defined as:

\[
\theta \sim \text{Uniform}(0.1, 0.9)^d.
\]

The sampling procedure is as follows: first, a binomial distribution is used to determine the number of samples from each component, with the expected number of uniform samples being \( n \cdot \delta \) and the expected number of Gaussian samples being \( n \cdot (1 - \delta) \), where \( n \) is the total number of samples. Samples are then drawn independently from each distribution and combined to form the final dataset.

\paragraph{Preprocessing} No preprocessing is required for this problem.

\paragraph{Hyperparameters}
\begin{itemize}
    \item $d = 5$: Dimensionality of the feature space.
    \item $\sigma$ = 0.02: Standard deviation of the Gaussian component.
    \item $\delta$ = 0.05: Weight of the uniform component in the mixture.
\end{itemize}

\subsection{Dirichlet-multinomial}

\paragraph{Sampling Process} The sampling model assumes that the parameter \(\theta\) is drawn from a Dirichlet distribution with a Gamma prior on the concentration parameters \(\alpha\). Specifically, we define \(\alpha\) as a vector of length \(K\), with each element sampled independently from a Gamma distribution: \(\alpha_k \sim \text{Gamma}(5, 0.5)\).

Given \(\alpha\), we sample \(\theta\) from a Dirichlet distribution \(\theta \sim \text{Dirichlet}(\alpha)\). The resulting \(\theta\) vector is used to draw \(n\) sets of multinomial counts $X$.

The observed counts $X$ are sampled from a multinomial distribution given \(\theta\) and a total sample size \(n_{\text{multi}} = 300\), i.e., \(X_i \sim \text{Multinomial}(n_{\text{multi}}, \theta)\), for \(i = 1, \ldots, n\). The counts are normalized by \(n_{\text{multi}}\) to provide relative frequencies.

\paragraph{Preprocessing} We subtract each element of $\theta$ by $\theta_K$, resulting in the output $\{\theta_i - \theta_K \mid i=1,2,\dots,K-1\}$. For the preprocessing of $X$, we input $X/n_{\text{multi}}$ into the neural network.

\paragraph{Hyperparameters}
\begin{itemize}
    \item \( K = 5 \) (Number of categories)
    \item \( n_{\text{multi}} = 300 \) (Total sample size for each multinomial draw)
    \item \( \alpha_i \sim \text{Gamma}(5, 0.5) \) $i=1,2\dots,K$
\end{itemize}

The resulting \(\theta\) is used in the multinomial distribution to generate observed counts.

\subsection{Poisson-gamma}

\paragraph{Sampling Process} The sampling model uses a Gamma distribution to generate \(\theta\), followed by a Poisson distribution to generate observed counts \(\mathbf{y}\). Specifically, each element of \(\theta_{ik}\) is sampled independently from a Gamma distribution: \(\theta_k \sim \text{Gamma}(\alpha, \beta)\), where \(\alpha\) is the shape parameter and \(\beta\) is the rate parameter.

Given these values of \(\theta\), observed counts \(\mathbf{y}\) are generated from a Poisson distribution, i.e., \(X_{ik} \sim \text{Poisson}(\theta_{ik})\), for \(i = 1, \ldots, n\) and \(k = 1, \ldots, \text{dim}(\theta)\).

\paragraph{Preprocessing} We input $\log(\theta_{ik})$ and $\log(X_{ik} + 1)$ as preprocessing.

\paragraph{Hyperparameters}
\begin{itemize}
    \item \( \alpha=2 \) (Shape parameter of the Gamma distribution)
    \item \( \beta=1 \) (Rate parameter of the Gamma distribution)
    \item $\text{dim}(\theta)=10$ (Dimension of the \(\theta\) vector)
\end{itemize}

\subsection{Socks}

This problem is motivated by the "Socks of Karl Broman" \href{https://www.sumsar.net/blog/2014/10/tiny-data-and-the-socks-of-karl-broman/}{Twitter post}: "That the first 11 socks are each distinct suggests that there are a lot more socks."  More generally: given that the first $k$ socks to come out of the laundry are unique, what is the posterior distribution over the total number of socks in the laundry?  

In the simple version of the problem, we place a prior over the total number of socks in Karl Broman’s laundry ($n_\text{socks}$) using a Negative Binomial distribution with mean 30 and standard deviation 15. The number of sock pairs ($n_\text{pairs}$) is determined by the proportion of socks in pairs ($\text{prop}\text{pairs}$), which follows a Beta distribution with shape parameters 15 and 2, with most of its mass between 0.75 and 1.0. The number of odd socks ($n\text{odd}$) is then calculated as the difference between the total socks and the paired socks.  The data consists of sampling socks one at a time, counting the number $k$ of socks that we draw before we find the first matched pair of socks.

To make the problem more interesting, we extend it to be a "multi-armed dryer", perhaps better known as "Karl Broman's laundromat," where each of $K$ dryers has its own distinct value of ($n_{\text{socks}}$), but where ($\text{prop}\text{pairs}$) is shared across all dryers.  We also introduce a more complex mixture prior for $\text{prop}\text{pairs}$.

\paragraph{Sampling Process} The sampling model utilizes a negative binomial distribution to generate the total number of socks in each dryer, followed by a simulation of randomly pulling socks until the first matching pair is found. The total number of socks across dryers is parameterized by a mean and standard deviation, which are converted into the negative binomial parameters \(r\) and \(p\) via the relationship \(r = \frac{\mu^2}{\sigma^2 - \mu}\) and \(p = 1 - \frac{\mu}{\sigma^2}\). Given these parameters, the total number of socks for each dryer is sampled: \(\text{total}_k \sim \text{NegBinom}(r, p)\), where \(k = 1, \ldots, K\).

The proportion of paired socks is modeled as a mixture of two beta distributions, with a common proportion sampled as \( \text{prop\_paired} \sim \text{BetaMix}(\alpha_1, \beta_1, \alpha_2, \beta_2, \text{mixing\_coefficient})\).

For each dryer, socks are randomly drawn until the first matching pair is found. This process generates a count of the number $X$ of draws required to observe the first match, and the count is normalized by the mean number of total socks.

\paragraph{Preprocessing} $total_k$ is normalized by the mean total socks, $\mu$; and $y$ is also normalized by $\mu$ before being input to the neural network.

\paragraph{Hyperparameters}
\begin{itemize}
    \item \( K = 10 \) (Number of dryers)
    \item \( \mu = 30 \) (Mean number of socks per dryer)
    \item \( \sigma = 15 \) (Standard deviation of the number of socks per dryer)
    \item \( \alpha_1 = 30, \beta_1 = 4 \) (Parameters for the first beta distribution)
    \item \( \alpha_2 = 50, \beta_2 = 50 \) (Parameters for the second beta distribution)
    \item \( \text{mixing\_coefficient} = 0.75 \) (Mixture proportion for the beta distributions)
    \item $K = 10$   Number of dryers

\end{itemize}

\subsection{Species Sampling}

\paragraph{Sampling Process} The sampling model assumes a Poisson distribution for the total number of birds observed in each survey, followed by multinomial sampling to allocate the birds to different species. The true proportions of species in the population are drawn from a Dirichlet distribution with a prior \(\alpha\), i.e., \(\mathbf{p} \sim \text{Dirichlet}(\alpha)\). For each survey, the total number of birds is sampled as \( N \sim \text{Poisson}(\lambda_{\text{total}}) \), and the species counts are distributed multinomially conditional on \(N\).

Observed counts are modeled with imperfect detection, where the detection probability for the first species follows a mixture model with two possible detection probabilities: \( p_{\text{easy}} = 0.9 \) and \( p_{\text{hard}} = 0.3 \). The detection probabilities for the other species are fixed: \( p_{\text{species 2}} = 0.6 \) and \( p_{\text{species 3}} = 0.7 \).

The observed counts are generated by applying a binomial model for detection, where the number of detected individuals for each species is drawn as \( X_{ij} \sim \text{Binomial}(N_{ij}, p_j) \), where \(N_{ij}\) is the true count of species \(j\) in survey \(i\), and \(p_j\) is the detection probability for species \(j\).

\paragraph{Preprocessing} $X_{ij}$ is normalized by $\lambda_{\text{total}}$, and $\bm{p}$ is adjusted by subtracting the baseline as $\{p_i - p_{n_{\text{species}}} \mid i=1,\dots,n_{\text{species}}-1\}$ before being input into the neural network.

\paragraph{Hyperparameters}
\begin{itemize}
    \item \( n_{\text{species}} = 3 \) (Number of species)
    \item \( \lambda_{\text{total}} = 50 \) (Average total number of birds per survey)
    \item \( \alpha_{\text{Dirichlet}} = 2.0 \) (Concentration parameter for Dirichlet prior)
    \item \( p_{\text{mixture}} = 0.7 \) (Mixture probability for species 1 detection)
    \item \( p_{\text{easy}} = 0.9 \), \( p_{\text{hard}} = 0.3 \) (Detection probabilities for species 1)
    \item \( p_{\text{species 2}} = 0.6 \), \( p_{\text{species 3}} = 0.7 \) (Fixed detection probabilities for species 2 and 3)
\end{itemize}

\subsection{Normal-Gamma}

\paragraph{Sampling Process} The sampling model assumes a normal distribution for the observed data \( X \), with parameters \(\mu\) and \(\sigma\) governing the mean and standard deviation. The variance \(\sigma^2\) is drawn from an inverse gamma prior, parameterized by shape \(d/2\) and scale \(2/\eta\), i.e., \(\sigma^2 \sim \text{InverseGamma}(d/2, 2/\eta)\).

The mean \(\mu\) is drawn from a normal distribution with mean zero and standard deviation \(\sigma/\sqrt{\kappa}\), i.e., \(\mu \sim \mathcal{N}(0, \sigma/\sqrt{\kappa})\).

Given these parameters, observed data \(X\) is sampled from a normal distribution, i.e., \(X_i \sim \mathcal{N}(\mu, \sigma)\), for \(i = 1, \ldots, n\).

\paragraph{Preprocessing} The log of the standard deviation, \(\log(\sigma)\), is used for numerical stability.

\paragraph{Hyperparameters:}
\begin{itemize}
    \item \( \mu = 0 \) (Prior mean of the normal distribution)
    \item \( \kappa = 1 \) (Precision parameter for the normal prior on \(\mu\))
    \item \( d = 8 \) (Shape parameter for the inverse gamma distribution)
    \item \( \eta = 8 \) (Scale parameter for the inverse gamma distribution)
\end{itemize}

\subsection{Normal-Wishart}

\paragraph{Sampling Process} The sampling model assumes a multivariate normal distribution for the observed data \( y \), with a Normal-Inverse-Wishart (NIW) prior on the mean vector \(\mu\) and covariance matrix \(\Sigma\). The NIW prior is parameterized by hyperparameters \(\mu_0\), \(\kappa_0\), \(\nu_0\), and \(\Psi_0\), where:
\(\mu \mid \Sigma \sim \mathcal{N}(\mu_0, \Sigma/\kappa_0)\),
\(\Sigma \sim \text{Inverse-Wishart}(\nu_0, \Psi_0)\).

Given the NIW prior, the parameters \(\mu\) and \(\Sigma\) are sampled, and data \( y \) is drawn from a multivariate normal distribution, \( X_i \sim \mathcal{N}(\mu, \Sigma) \), for \( i = 1, \ldots, n \).

\paragraph{Preprocessing} To avoid the positive-definite constraint of the covariance matrix, the covariance matrix $\Sigma$ is decomposed into its Cholesky components. The Cholesky factor is stored as the logarithm of its diagonal elements, while the off-diagonal elements are stored as is, ensuring that all elements input to the neural network lie in $\mathbb{R}$. No preprocessing needed for $X$.

\textbf{Hyperparameters:}
\begin{itemize}
    \item \( \mu_0 = \mathbf{0} \) (Prior mean vector)
    \item \( \kappa_0 = 1.0 \) (Scaling factor for the prior mean)
    \item \( \nu_0 = X_{\text{dim}} + 2 \) (Degrees of freedom for the Inverse-Wishart prior)
    \item \( \Psi_0 = I_{X_{\text{dim}}} \) (Scale matrix for the Inverse-Wishart prior)
    \item \( X_{\text{dim}} = 4 \) (Dimensionality of the multivariate normal)
\end{itemize}

\subsection{Multivariate G-and-K}

\paragraph{Sampling Process} The sampling model is based on the g-and-k distribution, parameterized by four variables: location \(a\), scale \(b\), skewness \(g\), and kurtosis \(k\). These parameters are drawn from their respective prior distributions:
\(a \sim \mathcal{N}(a_{\text{mean}}, a_{\text{std}})\);
\(b \sim \text{Gamma}(b_{\text{shape}}, b_{\text{scale}})\);
\(g \sim \mathcal{N}(g_{\text{mean}}, g_{\text{std}})\);
\(k \sim \text{Gamma}(k_{\text{shape}}, k_{\text{scale}})\).

The g-and-k distribution is defined through its quantile function, where the quantile \(z\) from the standard normal distribution is transformed using the parameters \(a\), \(b\), \(g\), and \(k\) as:
\[
Q(u; a, b, g, k) = a + b \cdot (1 + c \cdot \tanh\left(\frac{g z}{2}\right)) \cdot (1 + z^2)^k \cdot z,
\]
where \(u \sim \mathcal{U}(0, 1)\) and \(z = \Phi^{-1}(u)\) is the quantile of the standard normal distribution. Then we sample parameters $a,b,g,k$ from prior distribution and sample $X$ from $Q$ function.

\paragraph{Preprocessing} 

$a$ is normalized by $\text{scale\_odds\_order}$, and $b$ undergoes a log transformation before being normalized by $\text{scale\_even\_order}$. Similarly, $g$ is normalized by $\text{scale\_odds\_order}$, while $k$ is log-transformed and then scaled by $\text{scale\_even\_order}$. Finally, $X$ is transformed by $\tanh(X/\text{scale\_X})$ before being input into the neural network.

\paragraph{Hyperparameters}
\begin{itemize}
    \item \( a_{\text{mean}} = 0 \), \( a_{\text{std}} = 1 \) (Location parameter prior)
    \item \( b_{\text{shape}} = 5 \), \( b_{\text{scale}} = 1/5 \) (Scale parameter prior)
    \item \( g_{\text{mean}} = 0 \), \( g_{\text{std}} = 1 \) (Skewness parameter prior)
    \item \( k_{\text{shape}} = 7 \), \( k_{\text{scale}} = 1/7 \) (Kurtosis parameter prior)
    \item \( \text{scale\_odds\_order} = 5 \), \( \text{scale\_even\_order} = 2 \) (Scaling factors for \(a\), \(g\), \(b\), and \(k\))
    \item \( \text{scale\_X} = 10 \) (Scaling factor for observed data transformation)
\end{itemize}

\subsection{Lotka-Volterra}

\paragraph{Sampling Process} The sampling model simulates the Lotka-Volterra predator-prey system with added process and observation noise. The dynamics of the system are governed by the following set ordinary differential equations (ODEs):

\[
\frac{dX_1}{dt} = \alpha X_1 - \beta X_1 X_2, \quad \frac{dX_2}{dt} = -\gamma X_2 + \delta X_1 X_2
\]
where \(X_1\) and \(X_2\) represent the prey and predator populations, respectively, and the parameters \(\alpha\), \(\beta\), \(\gamma\), and \(\delta\) determine the interaction dynamics between the two species.

The observation noise is sampled from a Normal distribution with zero mean and covariance matrix $R$, and is added to $(X_1, X_2)$ to introduce uncertainty in the observed data without influencing the ODE itself.

\[
R = \begin{bmatrix}
\sigma_{r1}^2 & \text{covariance}_r \\
\text{covariance}_r & \sigma_{r2}^2
\end{bmatrix}
\]
with \(\text{covariance}_r = \rho_r \cdot \sigma_{r1} \cdot \sigma_{r2}\), where \(\rho_r\) controls the correlation between the observation noise of the two species.

\paragraph{Preprocessing} No preprocessing is applied for this problem.

\paragraph{Hyperparameters}
\begin{itemize}
    \item \( \alpha \sim \mathcal{U}(0.5, 1.0) \) (Growth rate of prey)
    \item \( \beta \sim \mathcal{U}(0.01, 0.1) \) (Interaction rate between prey and predator)
    \item \( \gamma \sim \mathcal{U}(0.01, 0.5) \) (Decay rate of predator)
    \item \( \delta \sim \mathcal{U}(0.005, 0.05) \) (Growth rate of predator due to prey consumption)
    \item \( \sigma_{r1} \sim \text{Gamma}(5, 1/5) \) (Prey observation noise standard deviation)
    \item \( \sigma_{r2} \sim \text{Gamma}(5, 1/5) \) (Predator observation noise standard deviation)
    \item \( \rho_r \sim \mathcal{U}(-0.1, 0.1) \) (Correlation between prey and predator observation noise)
\end{itemize}

The parameters \(\alpha\), \(\beta\), \(\gamma\), and \(\delta\) are used to simulate the predator-prey system, with observation noise added to the system state to generate the final observed data.

\subsection{Fractional Brownian motion.} 

\paragraph{Sampling Process} The sampling model involves simulating fractional Brownian motion (fBm) with a sinusoidal drift, where the drift term introduces periodic fluctuations. The prior distribution over the parameters \( \theta = (\text{hurst}, \tau^2, \text{amplitude}, \text{phase}, \text{period}) \) is defined by a combination of Beta, Gamma, and Uniform distributions, which are parameterized by the hyperparameters listed below. The parameter \( \text{hurst} \) controls the roughness of the fBm process, while \( \tau^2 \) governs the variance, and the sinusoidal drift is modulated by the amplitude, period, and phase. 

\paragraph{Preprocessing} We apply a log transformation to $\tau^2$ and the amplitude, and a logit transformation to the Hurst parameter. Additionally, the phase is transformed as $\text{phase} = \tan((\text{phase} - \pi)/2)$. No preprocessing is applied to the period or $X$.

\paragraph{Hyperparameters }
\begin{itemize}
    \item {hurst\_alpha} = 1.0, {hurst\_beta} = 1.0
    \item {tau2\_alpha} = 20.0, {tau2\_beta} = 1.0
    \item {amplitude\_alpha} = 3.0, {amplitude\_beta} = 0.2
    \item {phase\_min} = 0.0, {phase\_max} = 2$\pi$
\end{itemize}

\subsection{Stochastic volatility}

\paragraph{Sampling Process} The model assumes a univariate stochastic volatility process where the latent volatility \( h_t \) follows an autoregressive (AR(1)) model and the observed data \( y_t \) is generated based on this latent volatility. Specifically, the latent volatility \( h_t \) evolves according to:
\[
h_t = \mu + \phi (h_{t-1} - \mu) + \sigma_{\eta} \eta_t
\]
where \( \mu \) is the long-term mean of the volatility, \( \phi \) is the autoregressive coefficient, and \( \sigma_{\eta} \) is the volatility of volatility. The innovations \(\eta_t\) are normally distributed with mean zero and unit variance.

For the observed data, the model assumes:
\[
X_t = \exp\left( \frac{h_t}{2} \right) \epsilon_t
\]
where \(\epsilon_t \sim \mathcal{N}(0, 1)\) is the observation noise, and \( h_t \) determines the time-varying volatility of \( y_t \).

The latent volatility can be simulated using either an Euler-Maruyama approximation of the SDE or a recursive update loop for the AR(1) process.

\paragraph{Preprocessing} No preprocessing is applied for this problem.

\paragraph{Hyperparameters}
\begin{itemize}
    \item \( \mu \sim \mathcal{N}(-1, 0.5) \) (Long-term mean of the volatility)
    \item \( \phi \sim \mathcal{U}(0.9, 0.999) \) (Autoregressive coefficient, typically close to 1)
    \item \( \sigma_{\eta} \sim \mathcal{U}(0.1, 0.3) \) (Volatility of the latent volatility process)
\end{itemize}

The parameters \( \mu \), \( \phi \), and \( \sigma_{\eta} \) are used to generate the latent volatility process, which is then used to simulate the observed data \( X_t \).

\subsection{Markov-switching factor model}

\paragraph{Sampling Process} The Markov Switching Factor Model (MSFM) is a time series model where the observed data \( y_t \) is generated as a function of two latent states that switch according to a Markov chain. The latent factor loadings are modulated by the regime-dependent parameters \(\beta_1\) and \(\beta_2\), while the latent factor \( x_t \) follows an autoregressive process with noise.

For the Markov chain, the transition matrix is parameterized by \(\text{logit}(p_1)\) and \(\text{logit}(p_2)\), where the transition probabilities \(p_1\) and \(p_2\) govern the likelihood of remaining in states 1 and 2, respectively.

The observation noise \(\sigma_y\) and latent factor noise \(\sigma_x\) are sampled from Gamma distributions. Specifically, the observation noise is drawn as \(\sigma_y \sim \text{Gamma}(\sigma_{y,\text{shape}}, \sigma_{y,\text{scale}})\), and the latent factor noise is drawn similarly, \(\sigma_x \sim \text{Gamma}(\sigma_{x,\text{shape}}, \sigma_{x,\text{scale}})\).

The model switches between regimes at each time step based on the transition probabilities, and the observed data \( y_t \) is generated from a normal distribution with mean \( \beta_{\lambda_t} x_t \) and standard deviation \( \sigma_y \), where \(\lambda_t\) indicates the current regime.

\paragraph{Preprocessing} We apply a log transformation to $\sigma_x$ and $\sigma_y$, and a logit transformation to $p_1$ and $p_2$.

\textbf{Hyperparameters:}
\begin{itemize}
    \item \( \beta_1 \sim \mathcal{N}(\beta_{1,\text{mean}}, \beta_{1,\tau} \cdot \sigma_y) \) (Factor loading for regime 1)
    \item \( \beta_2 \sim \mathcal{N}(\beta_{2,\text{mean}}, \beta_{2,\tau} \cdot \sigma_y) \) (Factor loading for regime 2)
    \item \( \sigma_y \sim \text{Gamma}(\sigma_{y,\text{shape}}, \sigma_{y,\text{scale}}) \) (Observation noise standard deviation)
    \item \( \sigma_x \sim \text{Gamma}(\sigma_{x,\text{shape}}, \sigma_{x,\text{scale}}) \) (Latent factor noise standard deviation)
    \item \( p_1 \sim \text{Beta}(p_{1,a}, p_{1,b}) \) (Transition probability for remaining in regime 1)
    \item \( p_2 \sim \text{Beta}(p_{2,a}, p_{2,b}) \) (Transition probability for remaining in regime 2)
    \item \( \text{scale\_factor} = 1 \) (Scaling factor for the logits of transition probabilities)
\end{itemize}

The parameters \(\beta_1\), \(\beta_2\), \(\sigma_y\), and \(\sigma_x\) are used to generate the observed data \( y_t \), and the latent factor \( x_t \) is modeled as a random walk with noise \(\sigma_x\).

\subsection{VAR(P)}

\paragraph{Sampling Process} The model assumes a vector autoregressive (VAR) process for multivariate time series data, with parameters sampled from a Minnesota prior to ensure stationarity. The VAR process of order \( p \) for a multivariate time series \( X_t \) is defined as:

\[
X_t = \sum_{h=1}^{p} \Gamma_h X_{t-h} + \epsilon_t, \quad \epsilon_t \sim \mathcal{N}(0, \Sigma)
\]

where \( \Gamma_h \) are the lagged coefficient matrices, and \( \Sigma \) is the diagonal covariance matrix of the noise terms.

The parameters \(\Gamma_h\) are sampled from a Minnesota prior, where each element of the coefficient matrices is drawn from a normal distribution with mean \( \mu = \frac{\tau}{h} \) and standard deviation \( \sigma = \frac{\theta}{h} \), where \( \tau \) and \( \theta \) control the prior shrinkage, and \( h \) is the lag index.

To ensure stationarity, the sampled parameters \(\Gamma\) are shrunk until all eigenvalues of the corresponding companion matrix lie within the unit circle.

The noise covariance matrix \(\Sigma\) is sampled from an inverse gamma prior for the variances, i.e., \( \Sigma \sim \text{InverseGamma}(\sigma_{\text{shape}}, \sigma_{\text{scale}}) \).

\paragraph{Preprocessing} $\Sigma$ undergoes a log transformation and is then scaled by scale\_sigma, while $X$ is scaled by scale\_X before being input into the neural network.

\paragraph{Hyperparameters}
\begin{itemize}
    \item \( \tau = 0.45 \) (Shrinkage parameter for the Minnesota prior)
    \item \( \theta = 0.25 \) (Variance scaling parameter for the Minnesota prior)
    \item \( \sigma_{\text{shape}} = 2.0 \) (Shape parameter for the inverse gamma prior on the noise variances)
    \item \( \sigma_{\text{scale}} = \frac{1}{6.0} \) (Scale parameter for the inverse gamma prior on the noise variances)
    \item \( \text{scale\_sigma} = 3 \) (Scaling factor for the log of the noise variances)
    \item \( \text{scale\_X} = 100 \) (Scaling factor for the simulated data)
\end{itemize}

The parameters \(\Gamma\) and \(\Sigma\) are used to simulate the VAR time series data, where the lagged relationships and noise are generated based on the sampled values.

\newpage

\section{Evaluations for all benchmarks}
\label{app:evaluations}

In this section, we present additional evaluations of the benchmarks, including training loss and SBC validation as described in Appendix \ref{app:validation}.

\subsection{Loss evaluation}

We should note that in our setting, the loss conveys more information than just the convergence of training; it also serves as the test loss, since our data is generated from an infinite data stream as described in the sampling process of Appendix \ref{app:benchmark_problems}. This means that all data used for training and calculating the loss is new data, with no data revisiting. Thus, the loss is a good indicator that our model not only converges to a local minimum of the training loss, but is also capable of generating better posterior samples, as the loss represents (the upper bound of) the KL distance between the true posterior and the learned posterior (Proposition \ref{thm:diffusion_upperbound} and Proposition \ref{thm:summary_diffusion}).

However, the losses for different decoders are not directly comparable, as the loss for cNF calculates the exact KL divergence between approximate and true posteriors, whereas the diffusion loss calculates an upper bound of the KL divergence  (up to constants). Therefore, they are not mutually comparable, and we plot them in separate figures.

\begin{figure*}[ht]
    \centering
    \begin{subfigure}[c]{0.24\textwidth} % Set subfigure width to 32% of the text width
        \includegraphics[width=\textwidth]{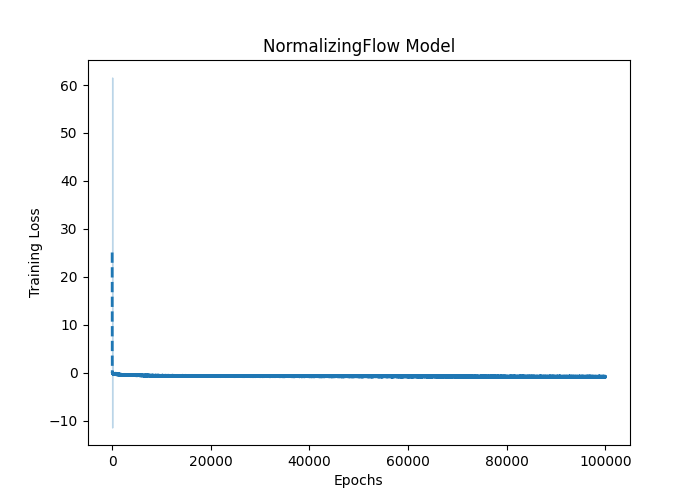}
        \caption{Sum of cosines: cNF}
    \end{subfigure}
    \hfill
    \begin{subfigure}[c]{0.24\textwidth} % Set subfigure width to 32% of the text width
        \includegraphics[width=\textwidth]{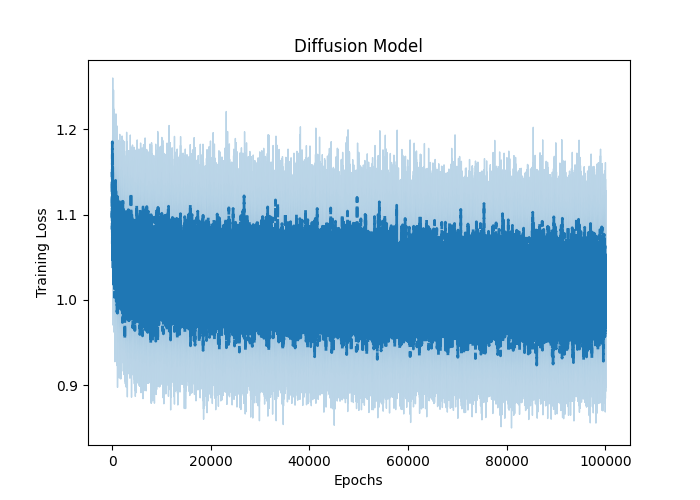}
        \caption{Sum of cosines: cDiff}
    \end{subfigure}
    \hfill
    \begin{subfigure}[c]{0.24\textwidth} % Set subfigure width to 32% of the text width
        \includegraphics[width=\textwidth]{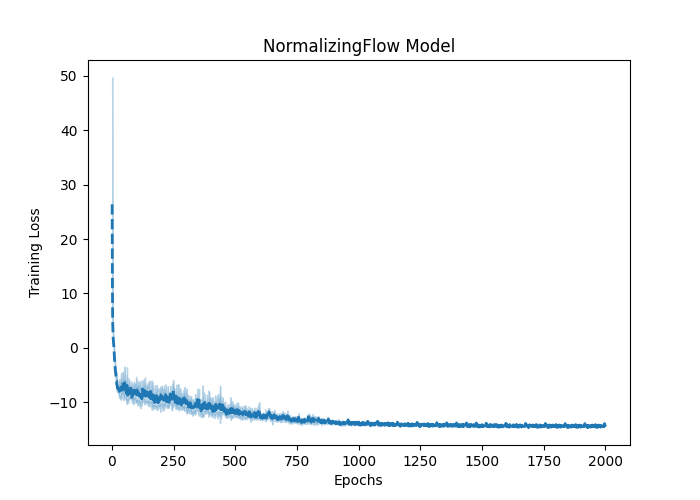}
        \caption{Witch’s hat: cNF}
    \end{subfigure}
    \begin{subfigure}[c]{0.24\textwidth} % Set subfigure width to 32% of the text width
        \includegraphics[width=\textwidth]{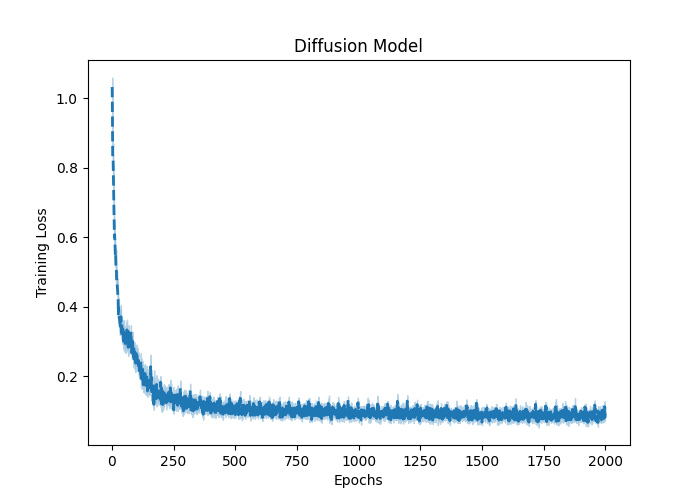}
        \caption{Witch’s hat: cDiff}
    \end{subfigure}
\end{figure*}

\begin{figure*}[ht]
    \centering
    \begin{subfigure}[c]{0.24\textwidth} % Set subfigure width to 32% of the text width
        \includegraphics[width=\textwidth]{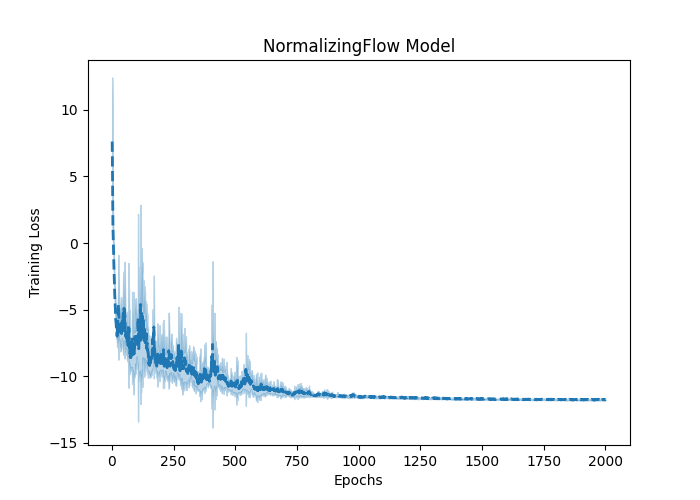}
        \caption{Dirichlet multinom: cNF}
    \end{subfigure}
    \hfill
    \begin{subfigure}[c]{0.24\textwidth} % Set subfigure width to 32% of the text width
        \includegraphics[width=\textwidth]{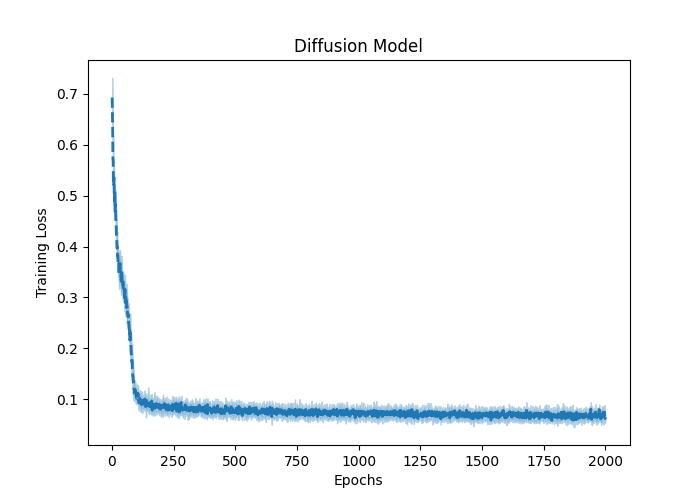}
        \caption{Dirichlet multinom: cDiff}
    \end{subfigure}
    \hfill
    \begin{subfigure}[c]{0.24\textwidth} % Set subfigure width to 32% of the text width
        \includegraphics[width=\textwidth]{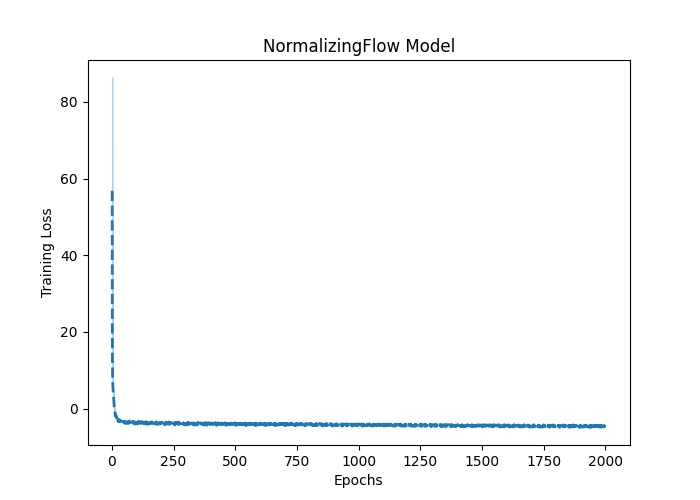}
        \caption{Socks: cNF}
    \end{subfigure}
    \begin{subfigure}[c]{0.24\textwidth} % Set subfigure width to 32% of the text width
        \includegraphics[width=\textwidth]{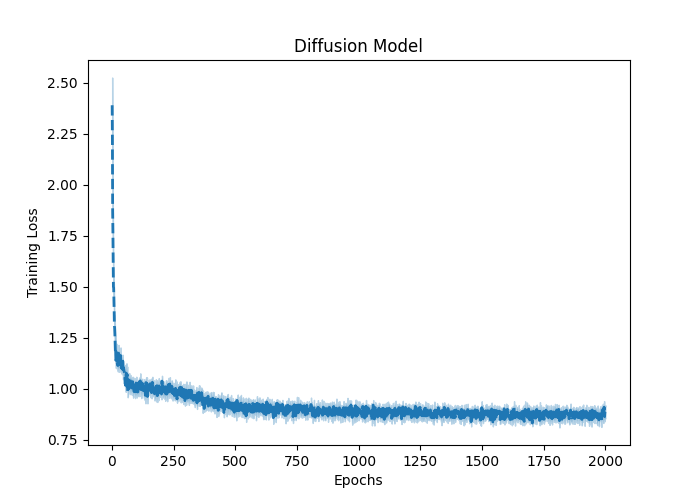}
        \caption{Socks: cDiff}
    \end{subfigure}
\end{figure*}

\begin{figure*}[ht]
    \centering
    \begin{subfigure}[c]{0.24\textwidth} % Set subfigure width to 32% of the text width
        \includegraphics[width=\textwidth]{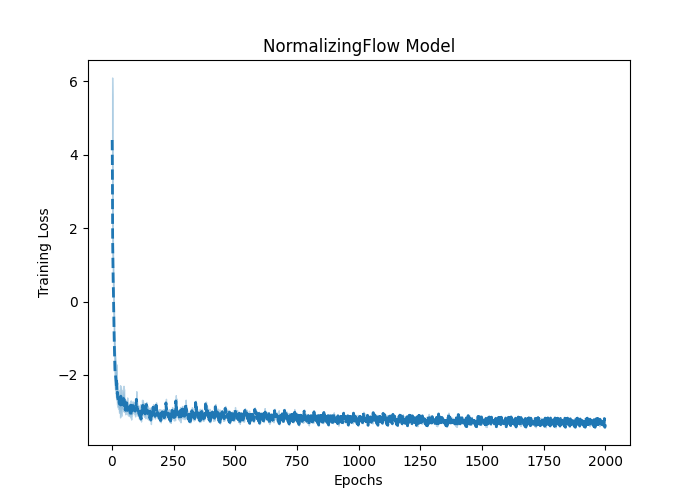}
        \caption{Species sampling: cNF}
    \end{subfigure}
    \hfill
    \begin{subfigure}[c]{0.24\textwidth} % Set subfigure width to 32% of the text width
        \includegraphics[width=\textwidth]{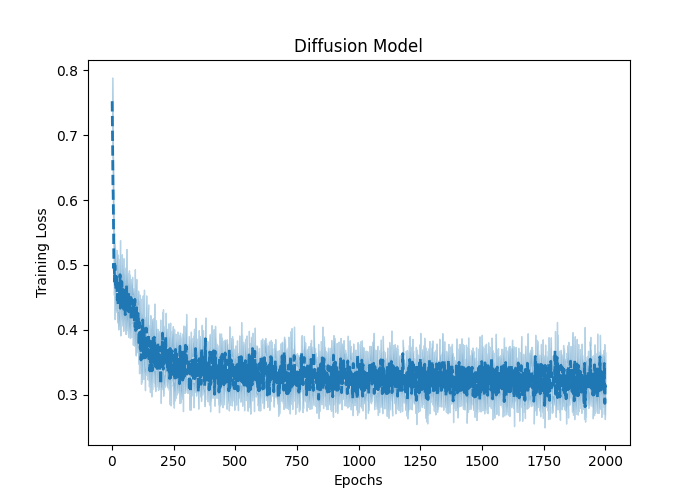}
        \caption{Species sampling: cDiff}
    \end{subfigure}
    \hfill
    \begin{subfigure}[c]{0.24\textwidth} % Set subfigure width to 32% of the text width
        \includegraphics[width=\textwidth]{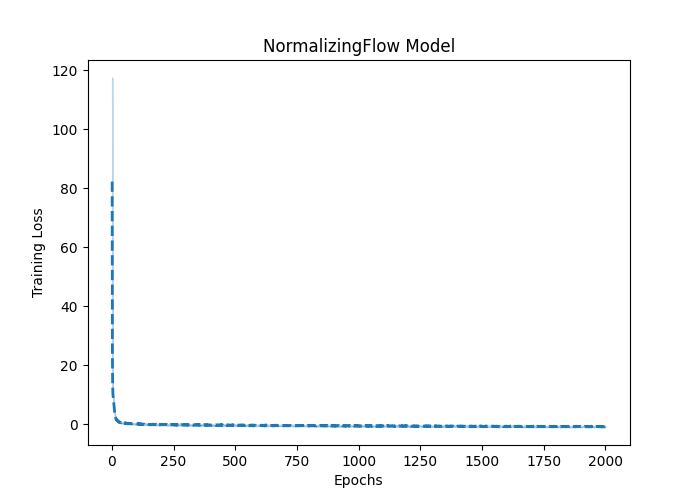}
        \caption{Poisson gamma: cNF}
    \end{subfigure}
    \begin{subfigure}[c]{0.24\textwidth} % Set subfigure width to 32% of the text width
        \includegraphics[width=\textwidth]{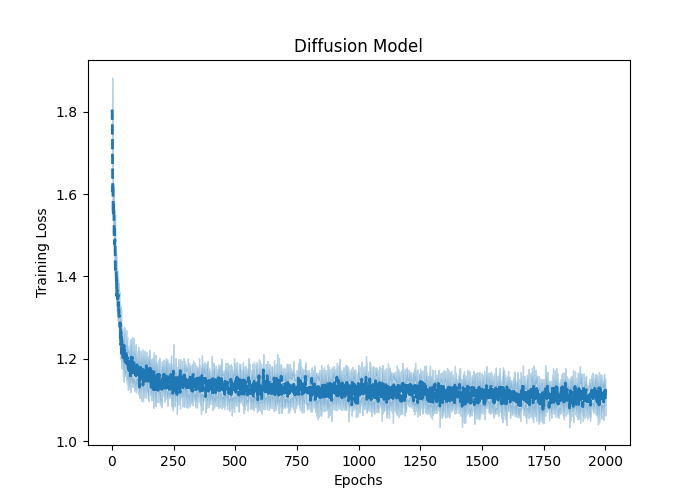}
        \caption{Poisson gamma: cDiff}
    \end{subfigure}
\end{figure*}

\begin{figure*}[ht]
    \centering
    \begin{subfigure}[c]{0.24\textwidth} % Set subfigure width to 32% of the text width
        \includegraphics[width=\textwidth]{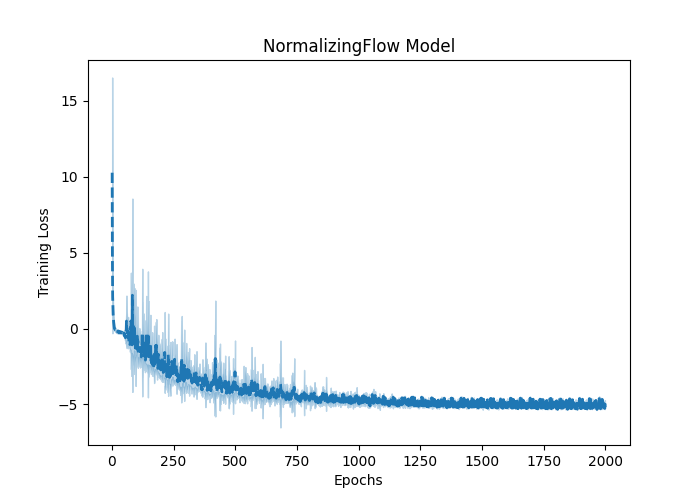}
        \caption{Normal gamma: cNF}
    \end{subfigure}
    \hfill
    \begin{subfigure}[c]{0.24\textwidth} % Set subfigure width to 32% of the text width
        \includegraphics[width=\textwidth]{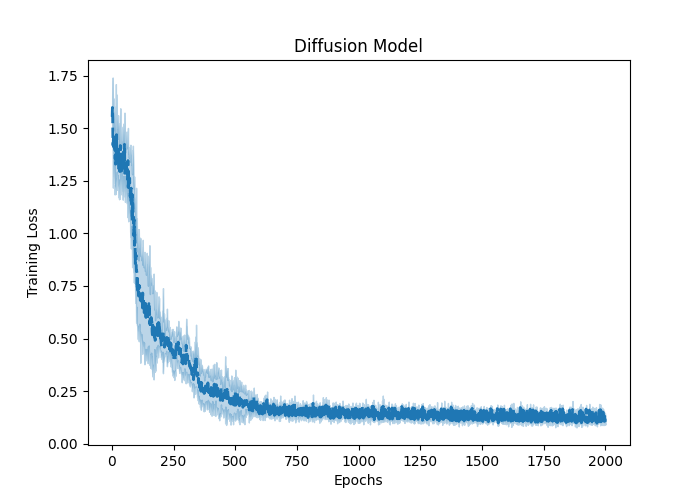}
        \caption{Normal gamma: cDiff}
    \end{subfigure}
    \hfill
    \begin{subfigure}[c]{0.24\textwidth} % Set subfigure width to 32% of the text width
        \includegraphics[width=\textwidth]{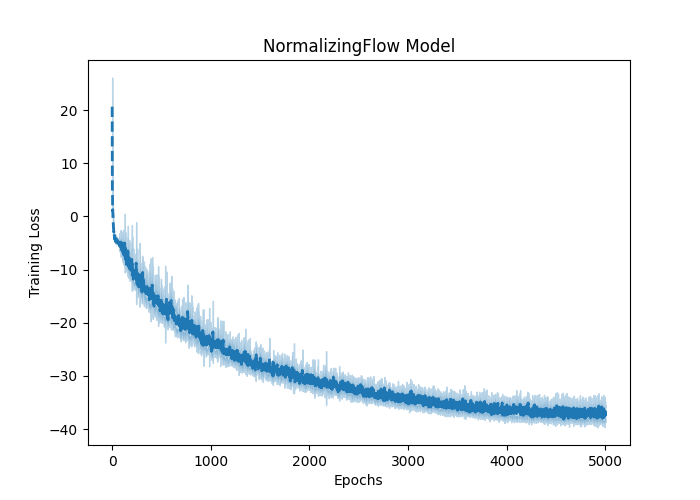}
        \caption{Normal wishart: cNF}
    \end{subfigure}
    \begin{subfigure}[c]{0.24\textwidth} % Set subfigure width to 32% of the text width
        \includegraphics[width=\textwidth]{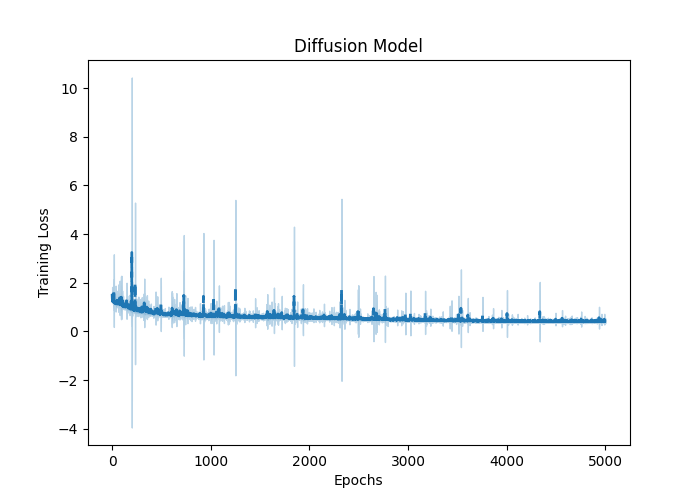}
        \caption{Normal wishart: cDiff}
    \end{subfigure}
\end{figure*}

\begin{figure*}[ht]
    \centering
    \begin{subfigure}[c]{0.24\textwidth} % Set subfigure width to 32% of the text width
        \includegraphics[width=\textwidth]{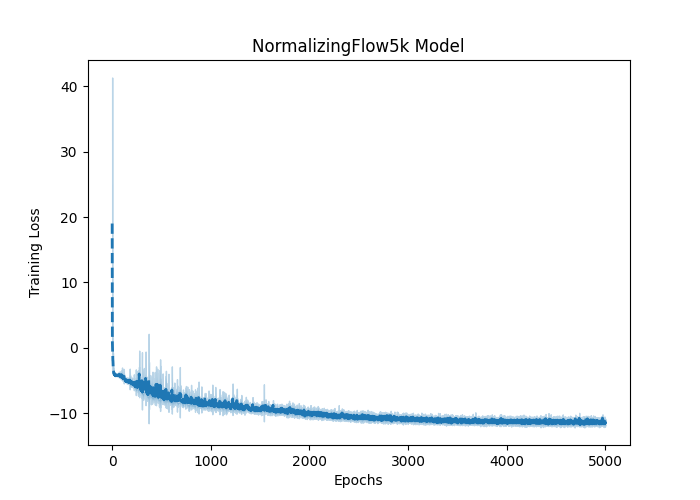}
        \caption{Multi G-and-K: cNF}
    \end{subfigure}
    \hfill
    \begin{subfigure}[c]{0.24\textwidth} % Set subfigure width to 32% of the text width
        \includegraphics[width=\textwidth]{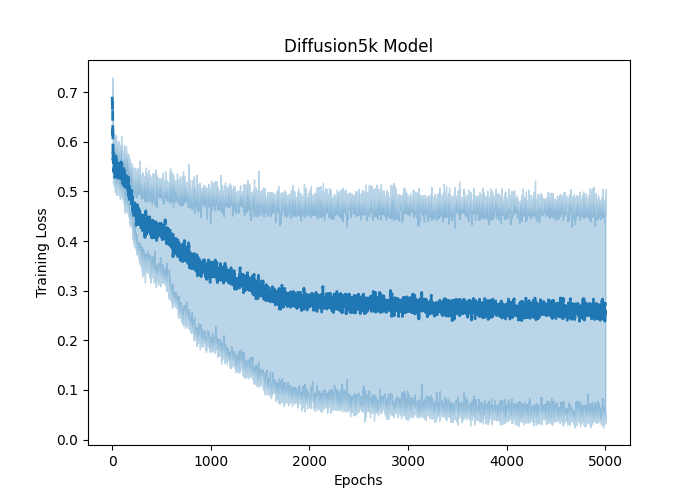}
        \caption{Multi G-and-K: cDiff}
    \end{subfigure}
    \hfill
    \begin{subfigure}[c]{0.24\textwidth} % Set subfigure width to 32% of the text width
        \includegraphics[width=\textwidth]{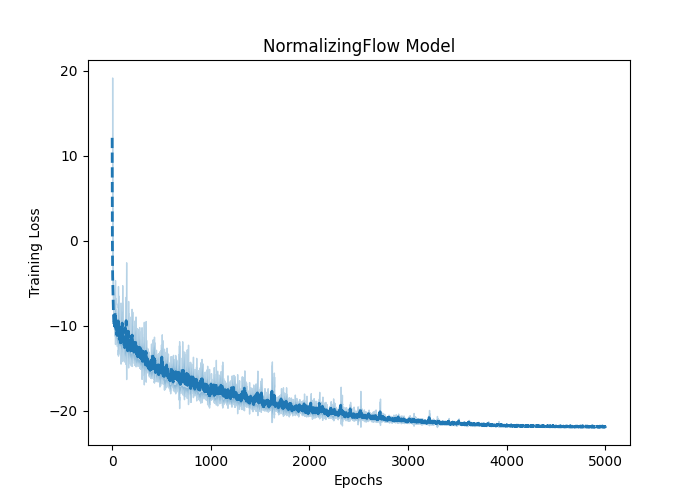}
        \caption{Lotka-Volterra: cNF}
    \end{subfigure}
    \begin{subfigure}[c]{0.24\textwidth} % Set subfigure width to 32% of the text width
        \includegraphics[width=\textwidth]{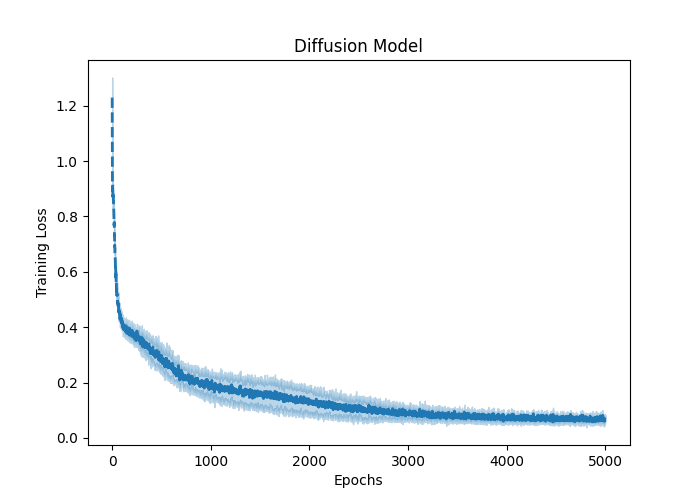}
        \caption{Lotka-Volterra: cDiff}
    \end{subfigure}
\end{figure*}

\begin{figure*}[ht]
    \centering
    \begin{subfigure}[c]{0.24\textwidth} % Set subfigure width to 32% of the text width
        \includegraphics[width=\textwidth]{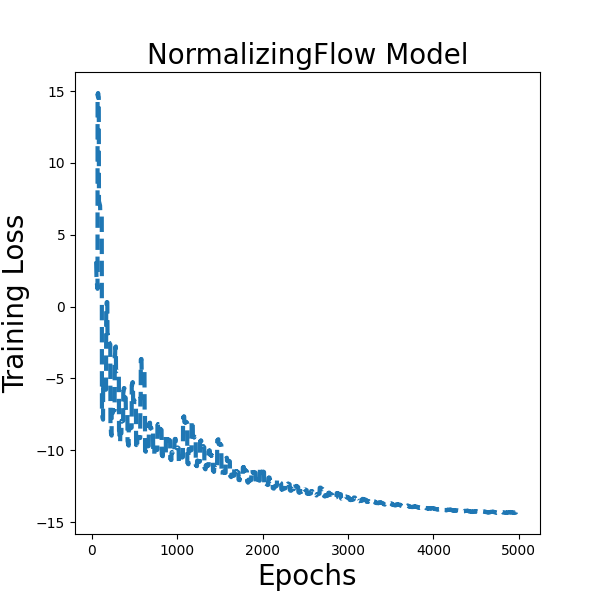}
        \caption{fractional BM: cNF}
    \end{subfigure}
    \hfill
    \begin{subfigure}[c]{0.24\textwidth} % Set subfigure width to 32% of the text width
        \includegraphics[width=\textwidth]{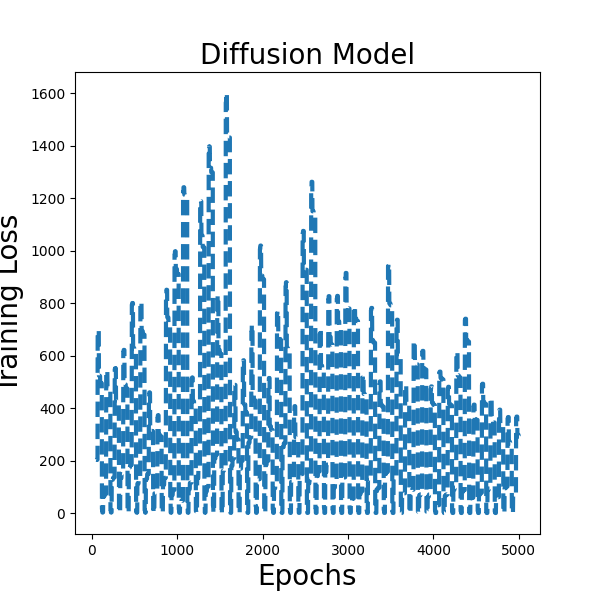}
        \caption{fractional BM: cDiff}
    \end{subfigure}
    \hfill
    \begin{subfigure}[c]{0.24\textwidth} % Set subfigure width to 32% of the text width
        \includegraphics[width=\textwidth]{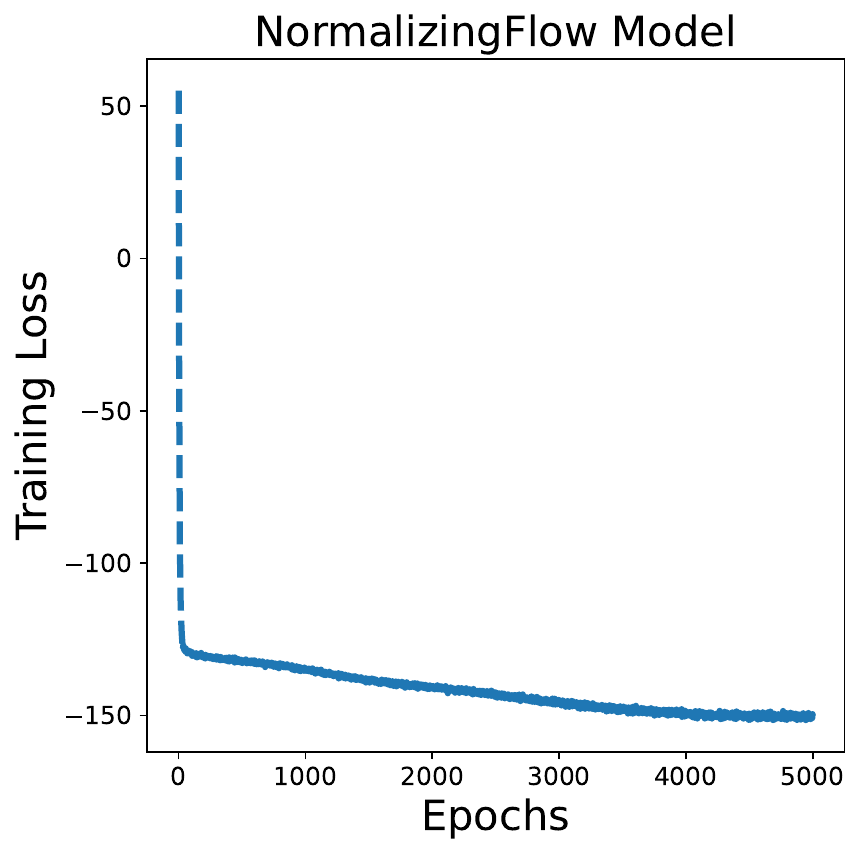}
        \caption{VAR(P): cNF}
    \end{subfigure}
    \begin{subfigure}[c]{0.24\textwidth} % Set subfigure width to 32% of the text width
        \includegraphics[width=\textwidth]{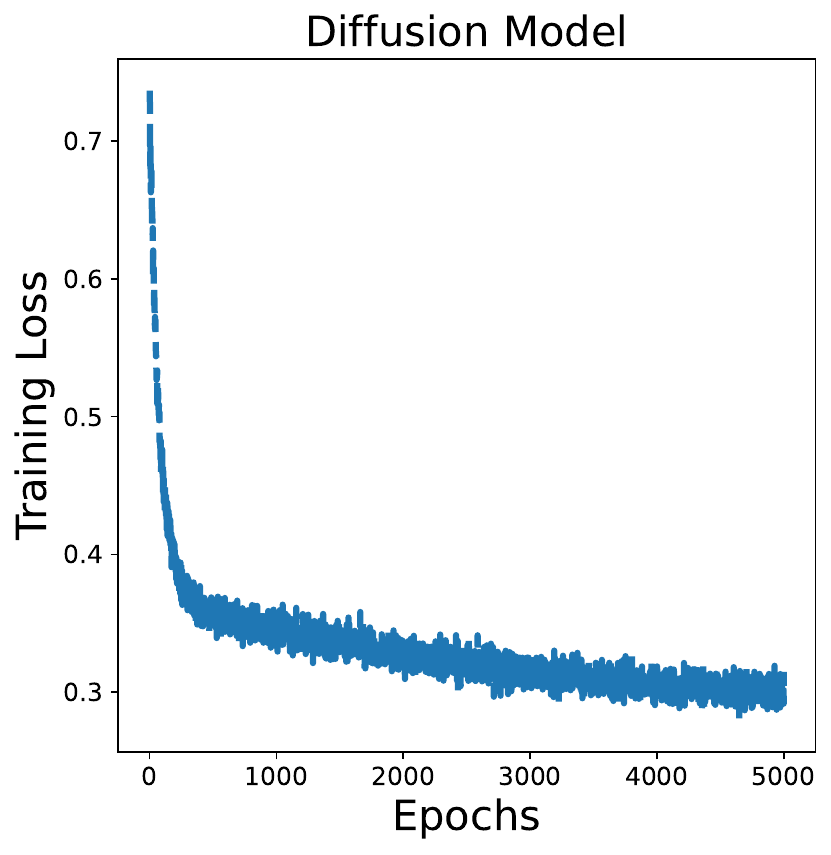}
        \caption{VAR(P): cDiff}
    \end{subfigure}
\end{figure*}

\begin{figure*}[ht]
    \centering
    \begin{subfigure}[c]{0.24\textwidth} % Set subfigure width to 32% of the text width
        \includegraphics[width=\textwidth]{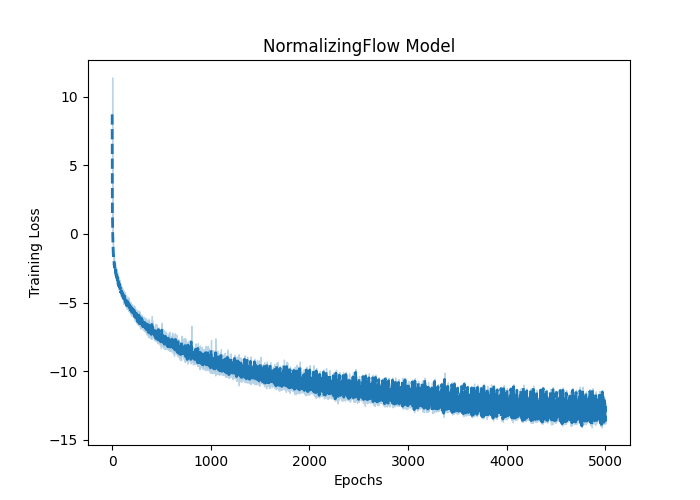}
        \caption{Markov switch: cNF}
    \end{subfigure}
    \hfill
    \begin{subfigure}[c]{0.24\textwidth} % Set subfigure width to 32% of the text width
        \includegraphics[width=\textwidth]{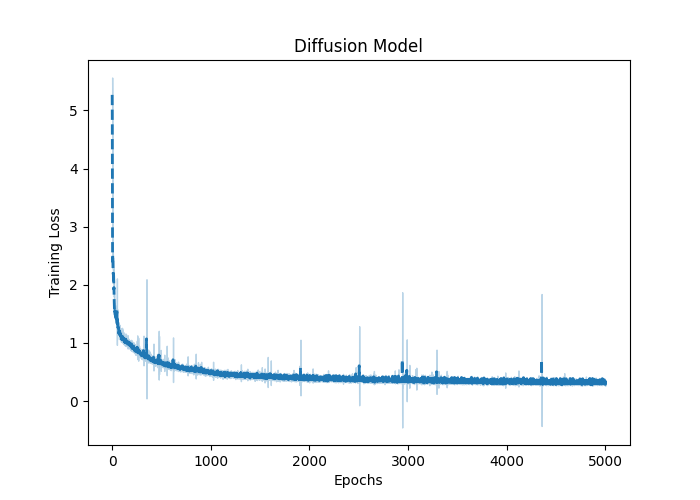}
        \caption{Markov switch: cDiff}
    \end{subfigure}
    \hfill
    \begin{subfigure}[c]{0.24\textwidth} % Set subfigure width to 32% of the text width
        \includegraphics[width=\textwidth]{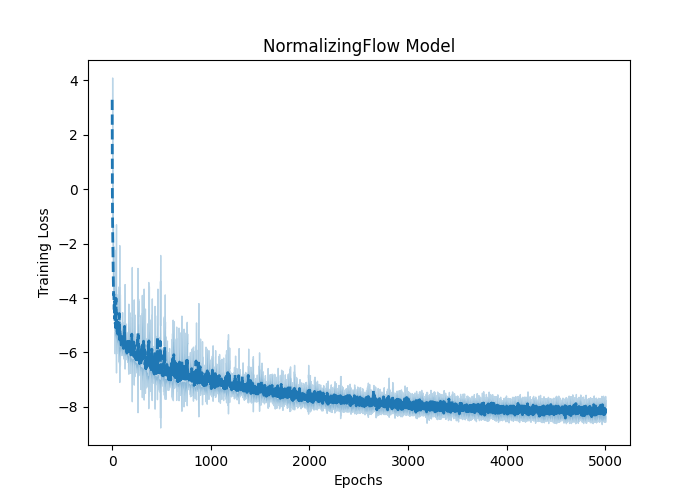}
        \caption{Stochastic vol: cNf}
    \end{subfigure}
    \begin{subfigure}[c]{0.24\textwidth} % Set subfigure width to 32% of the text width
        \includegraphics[width=\textwidth]{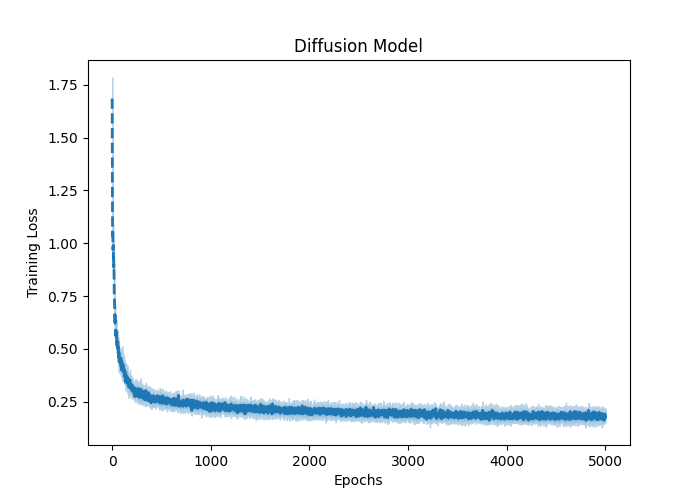}
        \caption{stochastic vol: cDiff}
    \end{subfigure}
\end{figure*}

\clearpage

\subsection{SBC evaluation}

We evaluate the distribution of SBC samples against a uniform distribution during the training process. Since the SBC distance is evaluated dimension by dimension, we plot the SBC distance versus epochs for the first dimension of $\theta$.

\begin{figure*}[ht]
    \centering
    \includegraphics[width=\linewidth]{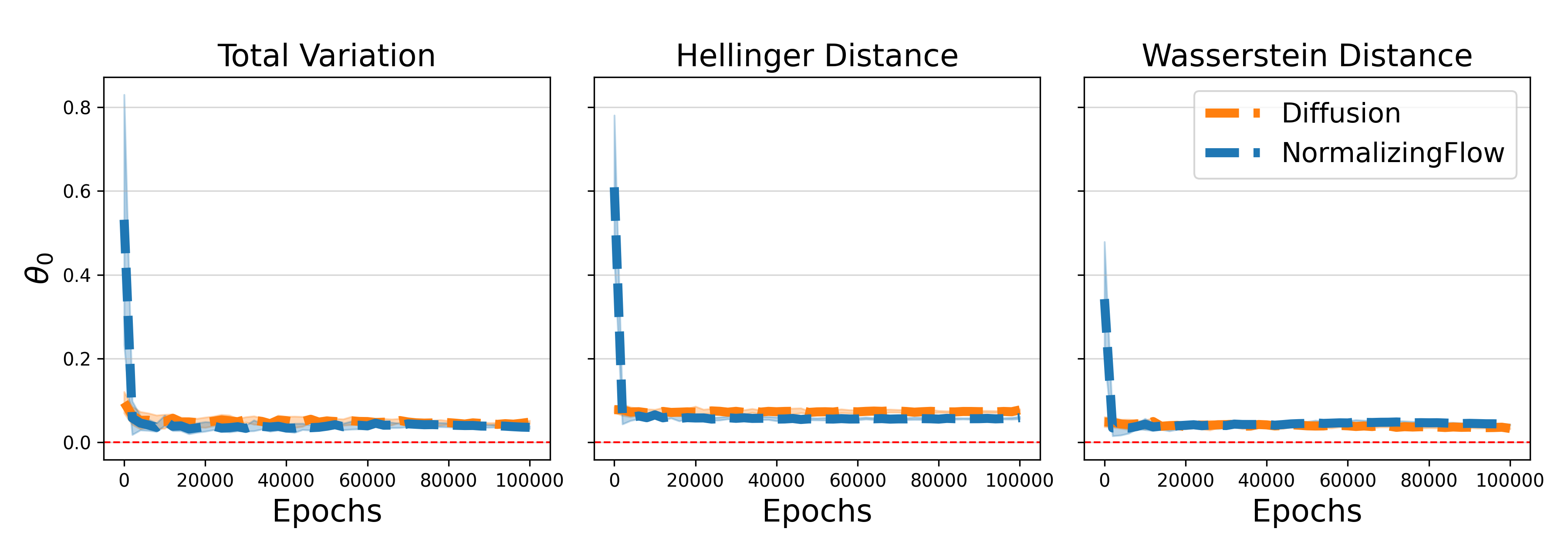}
    \caption{SBC distance versus epochs for the Sum of Cosines problem.}
\end{figure*}

\begin{figure*}[ht]
    \centering
    \includegraphics[width=\linewidth]{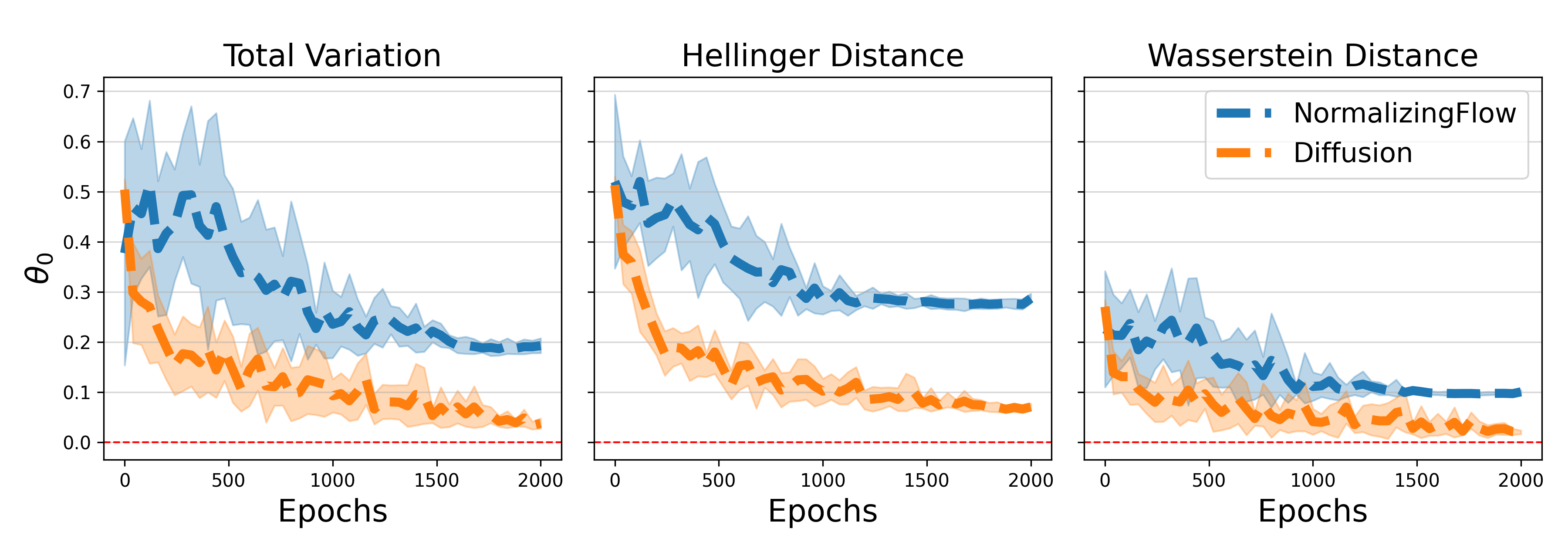}
    \caption{SBC distance versus epochs for the Witch's hat problem.}
\end{figure*}

\begin{figure*}[ht]
    \centering
    \includegraphics[width=\linewidth]{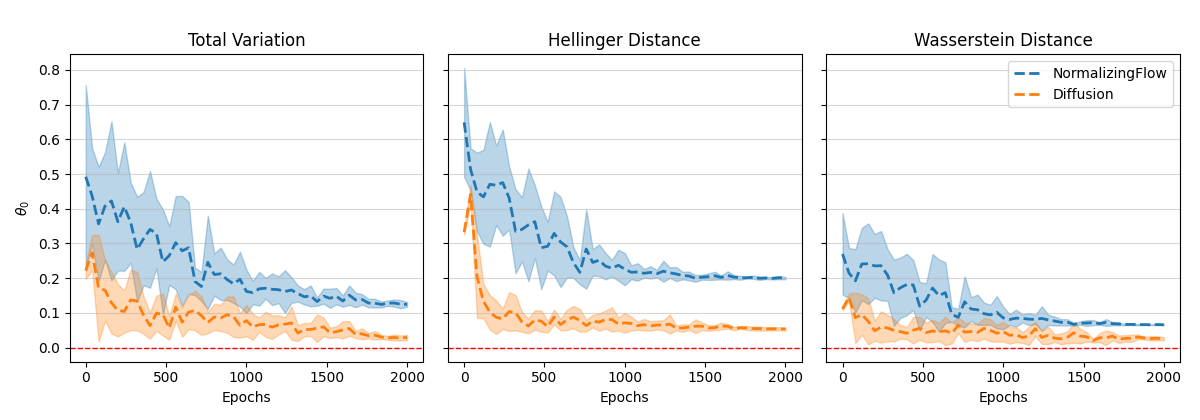}
    \caption{SBC distance versus epochs for the Dirichlet multinomial problem.}
\end{figure*}

\begin{figure*}[ht]
    \centering
    \includegraphics[width=\linewidth]{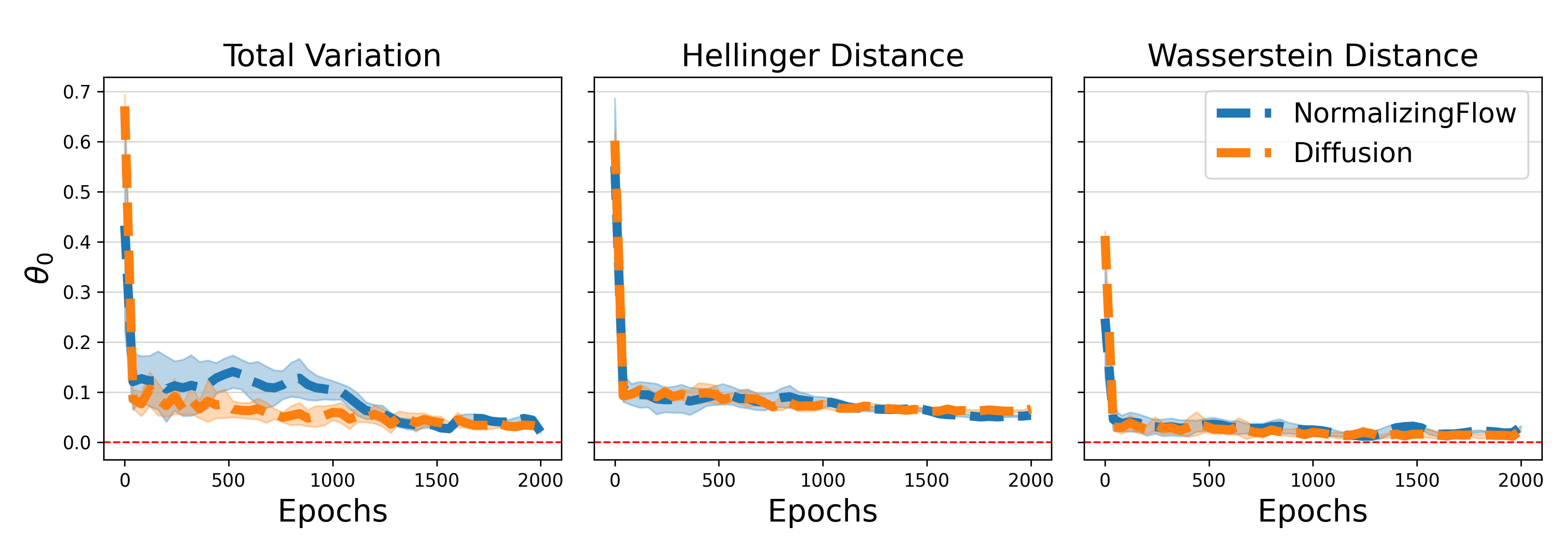}
    \caption{SBC distance versus epochs for the Socks problem.}
\end{figure*}

\begin{figure*}[ht]
    \centering
    \includegraphics[width=\linewidth]{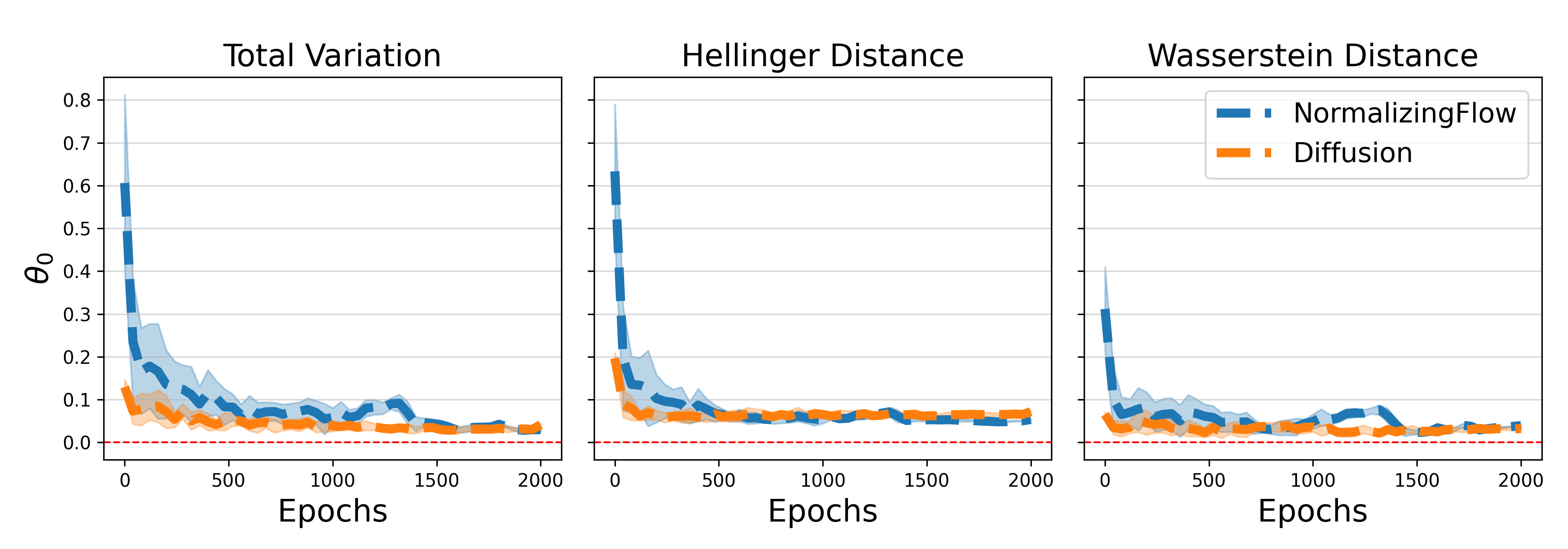}
    \caption{SBC distance versus epochs for the Species sampling problem.}
\end{figure*}

\begin{figure*}[ht]
    \centering
    \includegraphics[width=\linewidth]{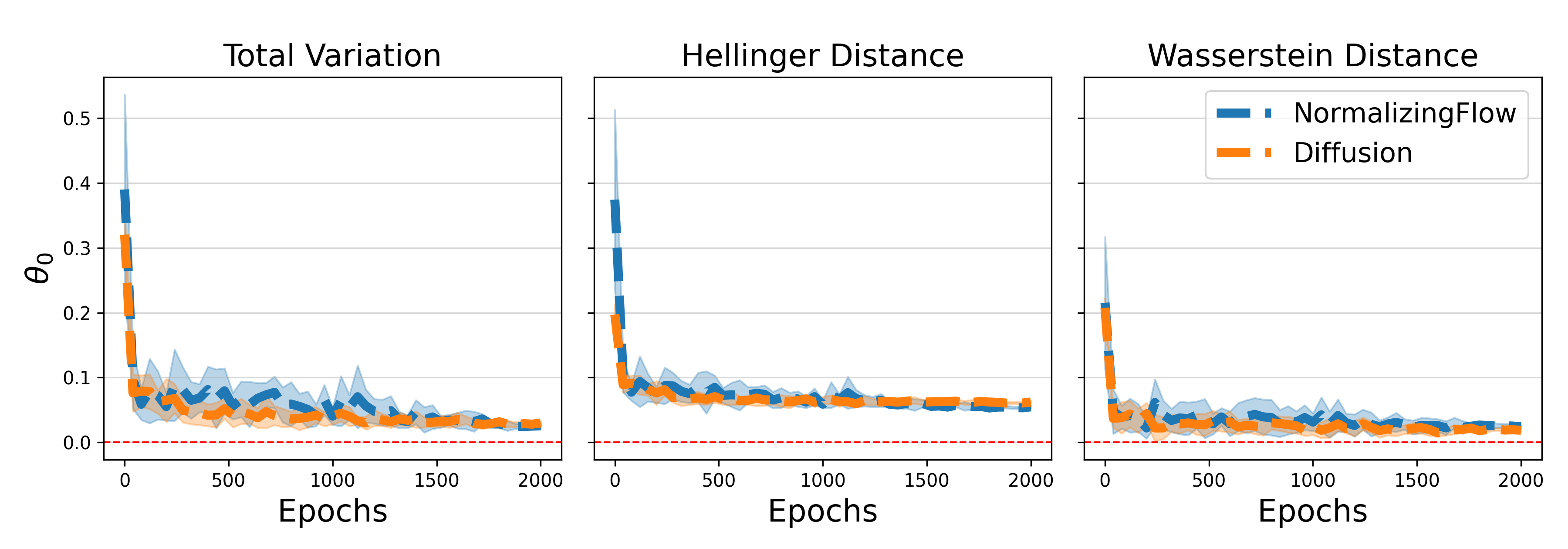}
    \caption{SBC distance versus epochs for the Possion gamma problem.}
\end{figure*}

\begin{figure*}[ht]
    \centering
    \includegraphics[width=\linewidth]{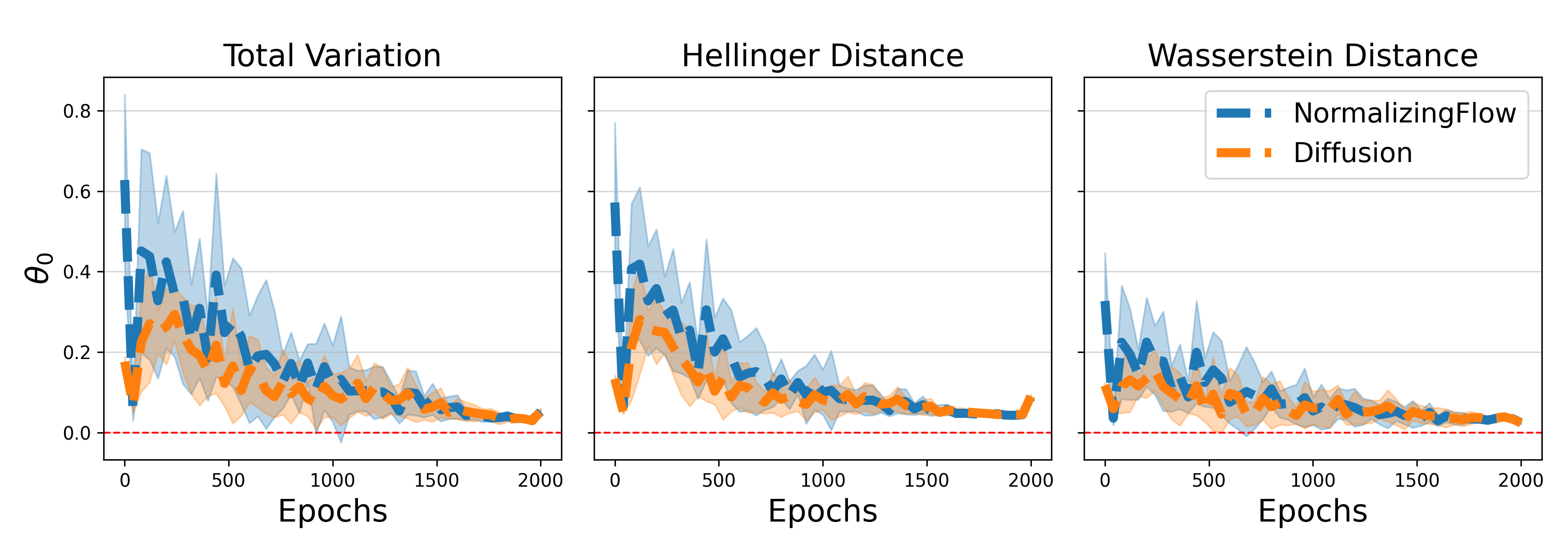}
    \caption{SBC distance versus epochs for the Normal gamma problem.}
\end{figure*}

\begin{figure*}[ht]
    \centering
    \includegraphics[width=\linewidth]{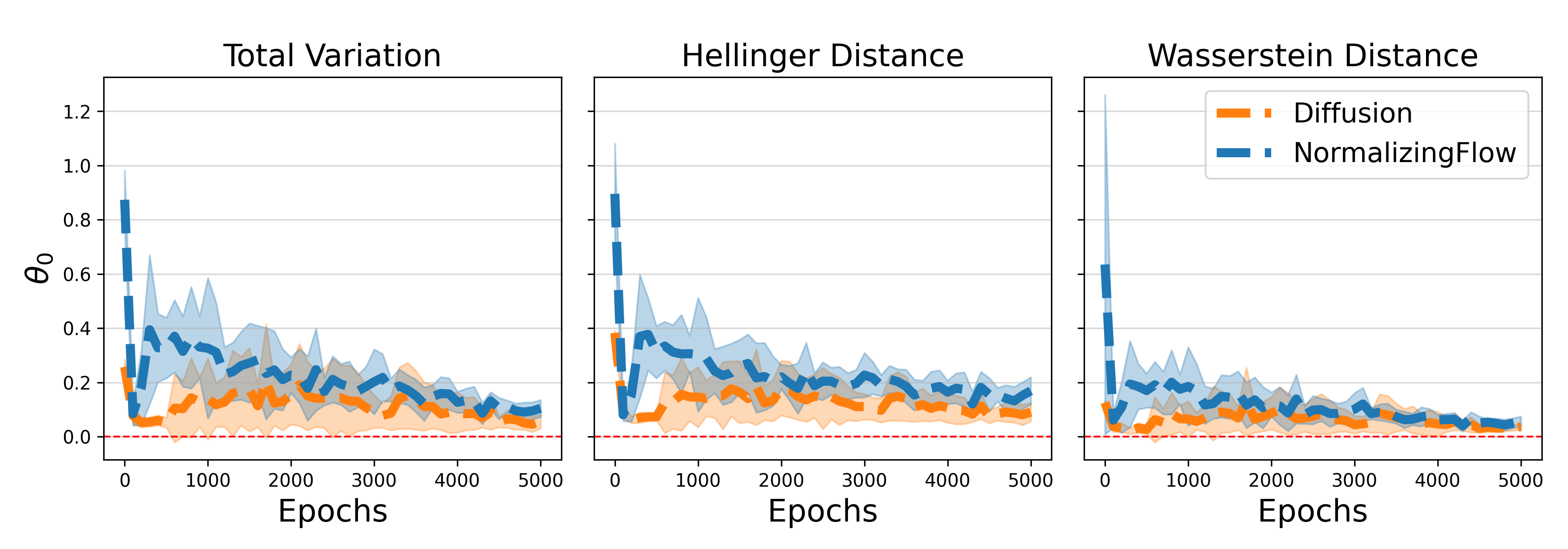}
    \caption{SBC distance versus epochs for the Multivariate g-and-k problem.}
\end{figure*}

\begin{figure*}[ht]
    \centering
    \includegraphics[width=\linewidth]{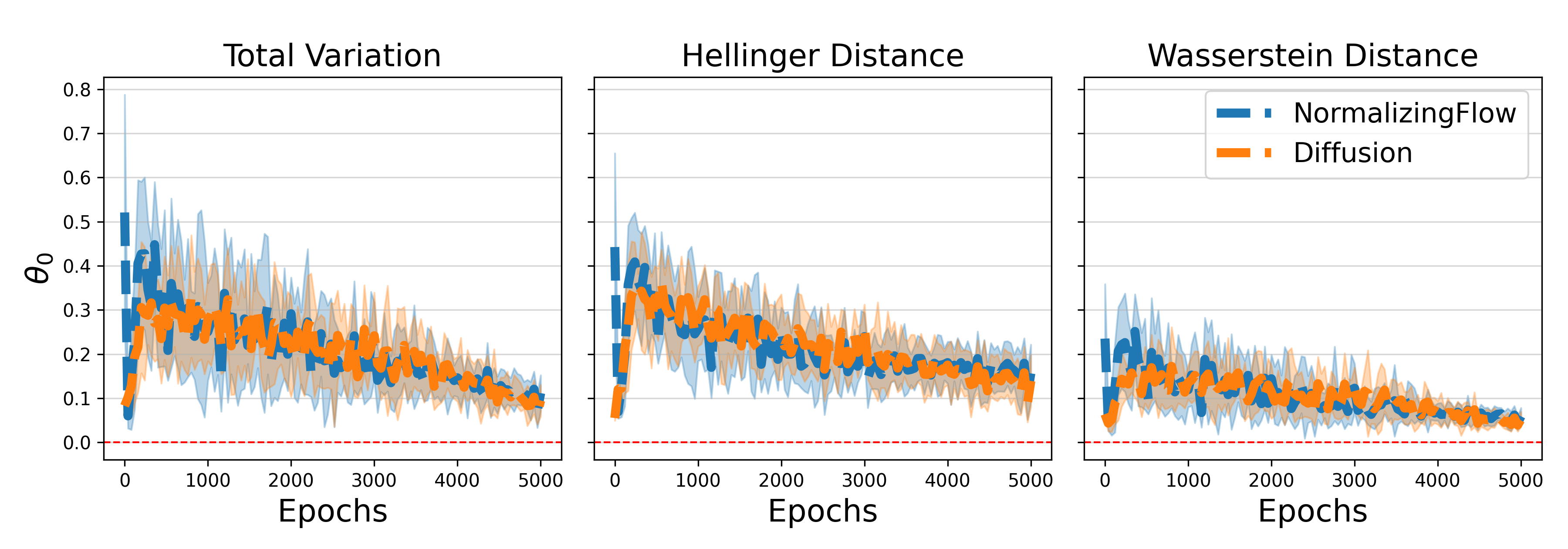}
    \caption{SBC distance versus epochs for the Normal Wishart problem.}
\end{figure*}

\begin{figure*}[ht]
    \centering
    \includegraphics[width=\linewidth]{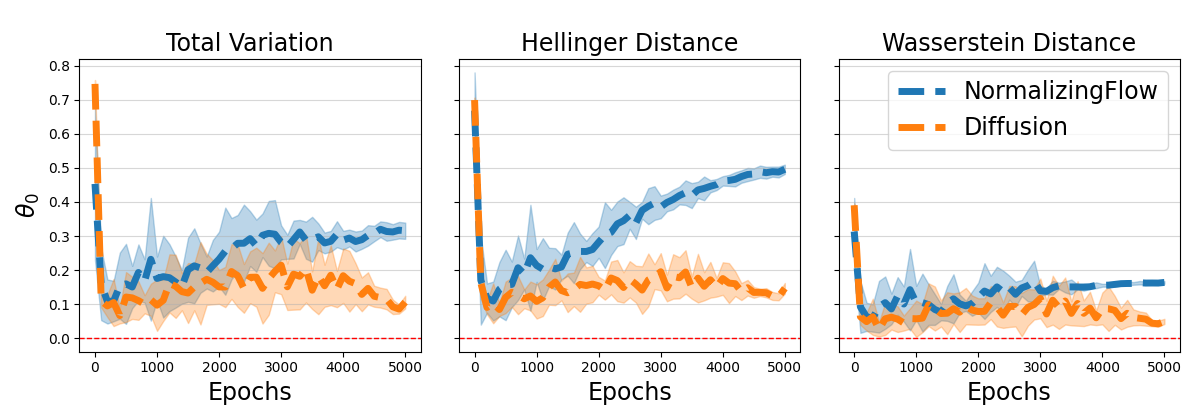}
    \caption{SBC distance versus epochs for the Lotka–Volterra problem.}
\end{figure*}

\begin{figure*}[ht]
    \centering
    \includegraphics[width=\linewidth]{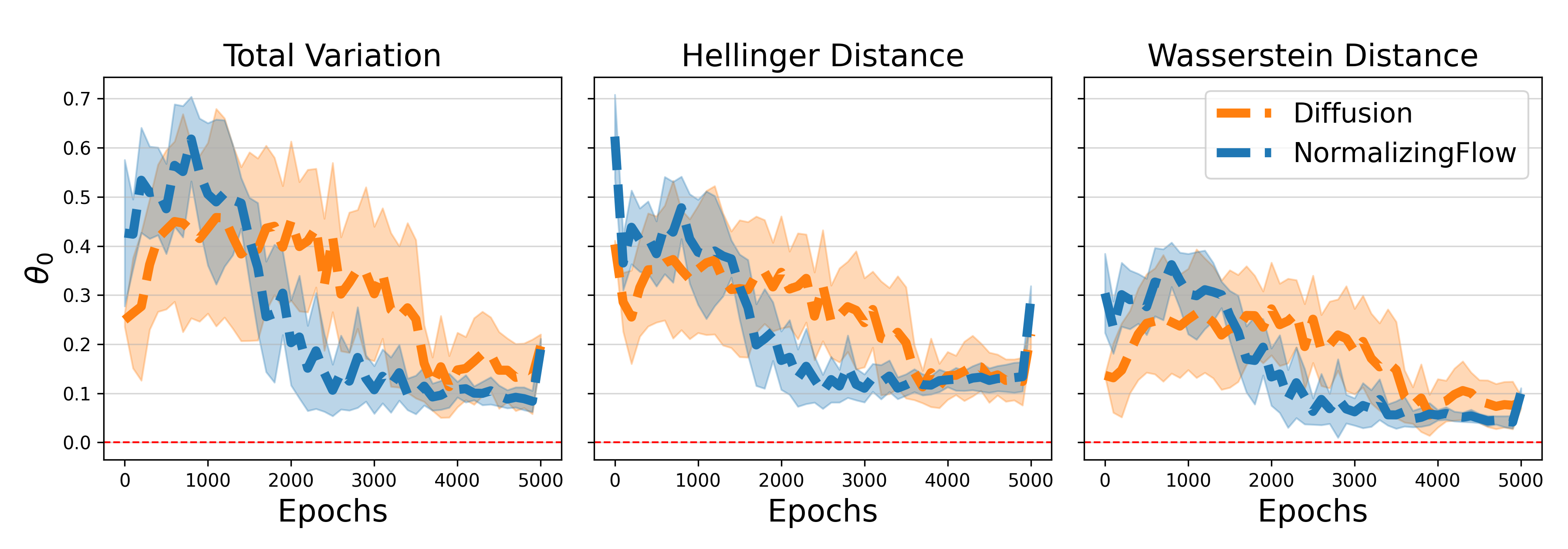}
    \caption{SBC distance versus epochs for the fractional BM problem.}
\end{figure*}

\begin{figure*}[ht]
    \centering
    \includegraphics[width=\linewidth]{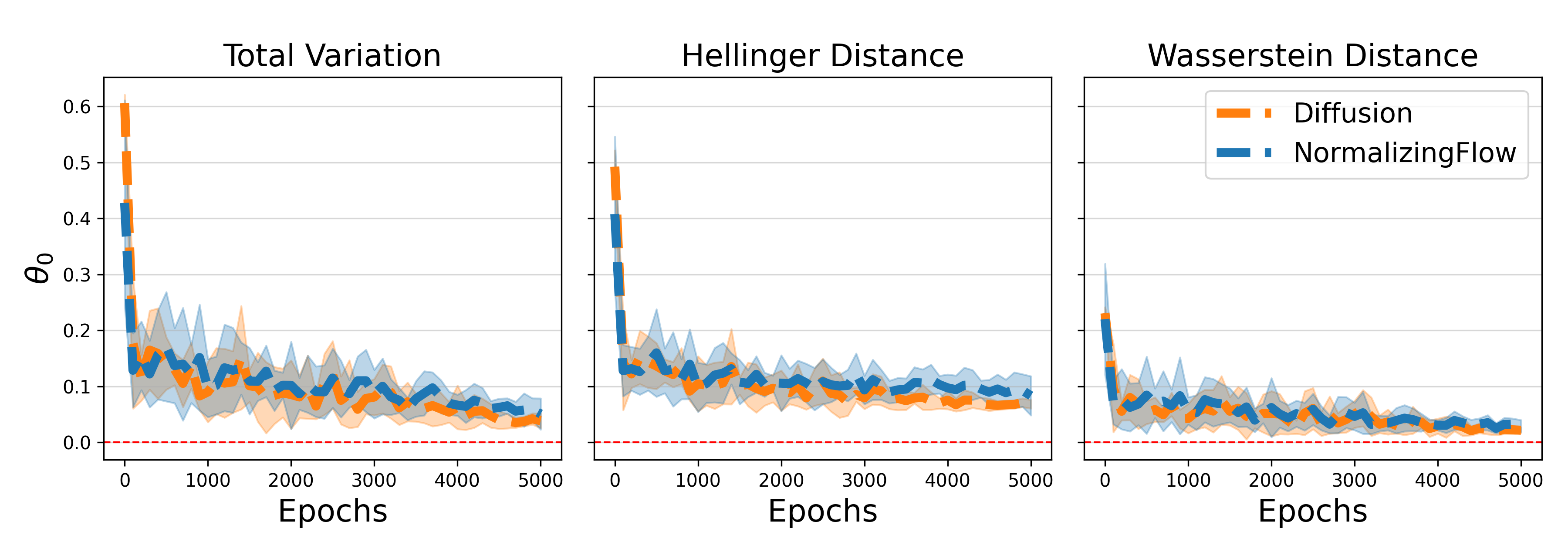}
    \caption{SBC distance versus epochs for the stochastic vol problem.}
\end{figure*}

\begin{figure*}[ht]
    \centering
    \includegraphics[width=\linewidth]{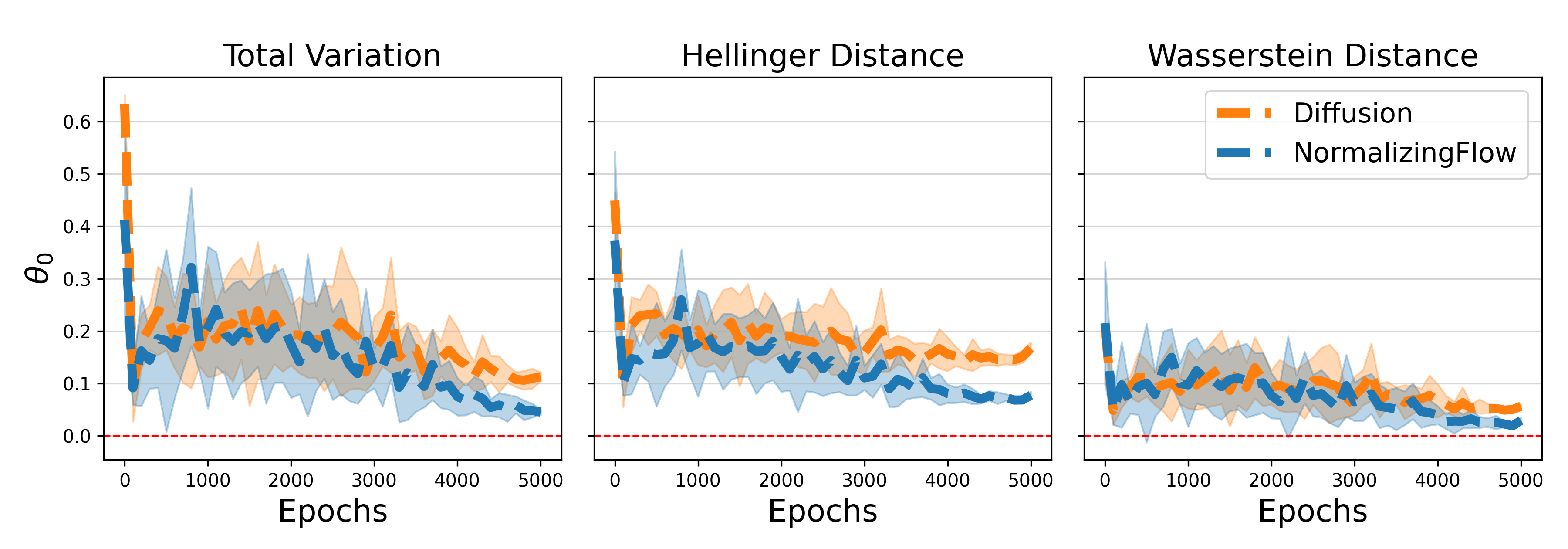}
    \caption{SBC distance versus epochs for the Markov switch problem.}
\end{figure*}

\begin{figure*}[ht]
    \centering
    \includegraphics[width=\linewidth]{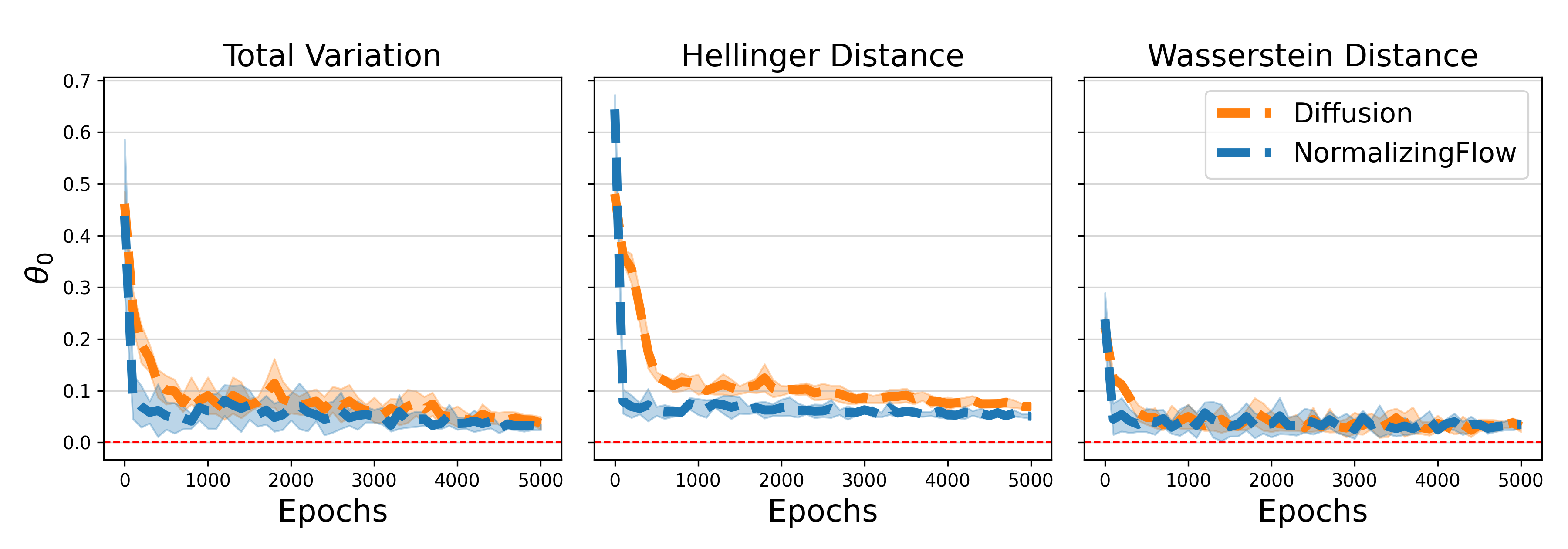}
    \caption{SBC distance versus epochs for the VAR(P) problem.}
\end{figure*}

%\end{document}